\documentclass[10pt,journal]{IEEEtran}

\usepackage{mathrsfs}

\usepackage{latexsym}
\usepackage{amssymb}
\usepackage{graphicx}
\usepackage{amsmath}
\usepackage{epsfig}
\usepackage{latexsym}
\usepackage{indentfirst}
\usepackage{amsthm}
\usepackage{subfigure}
\usepackage{amsfonts}
\usepackage{picins}

\ifCLASSOPTIONcompsoc
\else
\fi
\ifCLASSINFOpdf
\else
\fi

\usepackage{graphicx}

\usepackage{multirow}
\newtheorem{theorem}{Theorem}
\newtheorem{assumption}{Assumption}

\newtheorem{lemma}{Lemma}

\newtheorem{proposition}{Proposition}

\graphicspath{{figures/}}

\begin{document}

\title{Realization of spatial sparseness by  deep ReLU nets with massive data}

\author{Charles K. Chui,  Shao-Bo Lin, Bo Zhang, and Ding-Xuan Zhou
\IEEEcompsocitemizethanks{\IEEEcompsocthanksitem C. K. Chui and Bo Zhang are with
Department of Mathematics, Hong Kong Baptist University. C.K. Chui is also associated with the
Department of Statistics,  Stanford University,
CA 94305, USA. Shao-Bo Lin  is with the Center of Intelligent Decision-making and  Machine Learning,  School of Management, Xi'an Jiaotong  University, Xi'an, China.  D. X. Zhou is with
School of Data Science and Department of Mathematics, City University of Hong Kong, {  Hong Kong}. The corresponding author is S. B. Lin (email:
sblin1983@gmail.com).}}

  \IEEEcompsoctitleabstractindextext{

\begin{abstract}
The great success of deep learning poses urgent challenges for understanding its working mechanism and rationality. The depth, structure, and massive size of the data are recognized to be three key ingredients for deep learning. Most of the recent theoretical studies for deep learning focus on the necessity and advantages of depth and structures of neural networks. In this paper, we aim at rigorous verification of the importance of massive data in embodying the out-performance of deep learning. To approximate and learn spatially sparse and smooth functions, we establish a novel sampling theorem in learning theory to show the necessity of massive data. We then prove that implementing the classical empirical risk minimization on some deep nets facilitates in realization of the optimal learning rates derived in the sampling theorem. This perhaps explains why deep learning performs so well in the era of big data.

\end{abstract}

}

\maketitle

\IEEEdisplaynotcompsoctitleabstractindextext

\IEEEpeerreviewmaketitle


\section{Introduction}
With the rapid development of data mining and knowledge discovery, data of massive size are collected in  various disciplines \cite{Zhou2014}, including  medical
diagnosis, financial market analysis,  computer vision,
 natural language processing, time series forecasting, and search
engines.
These massive data  bring additional opportunities to discover  subtle data features which cannot be reflected by data of small size while creating a crucial challenge  on machine learning to develop learning schemes to realize  benefits by exploring the use of massive data. Although   numerous  learning schemes  such as distributed learning \cite{Lin2017JMLR}, localized learning \cite{Meister2016} and sub-sampling \cite{Gittens2016} have been proposed to handle massive data, all these schemes focused on the tractability   rather than the benefit  of massiveness. Therefore, it remains open to explore the benefits brought from massive data and to develop feasible learning strategies for realizing these benefits.

Deep learning \cite{Hinton2006}, characterized by training deep neural networks (deep nets for short) to extract data features by using  rich computational resources such as  computational power of modern graphical processor units (GPUs) and custom processors, has made  remarkable success in  computer vision \cite{Krizhevsky2012}, speech
recognition \cite{Lee2009} and game theory  \cite{Silver2016},
practically showing its power  in tackling massive data.
Recent developments on   deep learning theory also  provide several  exciting
theoretical results to explain the efficiency and rationality of deep learning.
In particular, numerous data features such as   manifold structures of the input space \cite{Shaham2015}, piecewise smoothness \cite{Petersen2017}, rotation-invariance \cite{Chui2018a} and sparseness in the frequency domain \cite{Schwab2018} were proved to be realizable by deep nets but cannot be extracted by shallow neural networks (shallow nets for short) with same order of free parameters.
All these interesting  studies theoretically  verify the necessity of depth in deep learning. The problem is, however, they do not provide any explanations on why deep learning works so well for big data.

\begin{figure}[!t]
\begin{minipage}[b]{0.49\linewidth}
\centering
\includegraphics*[width=4cm,height=3cm]{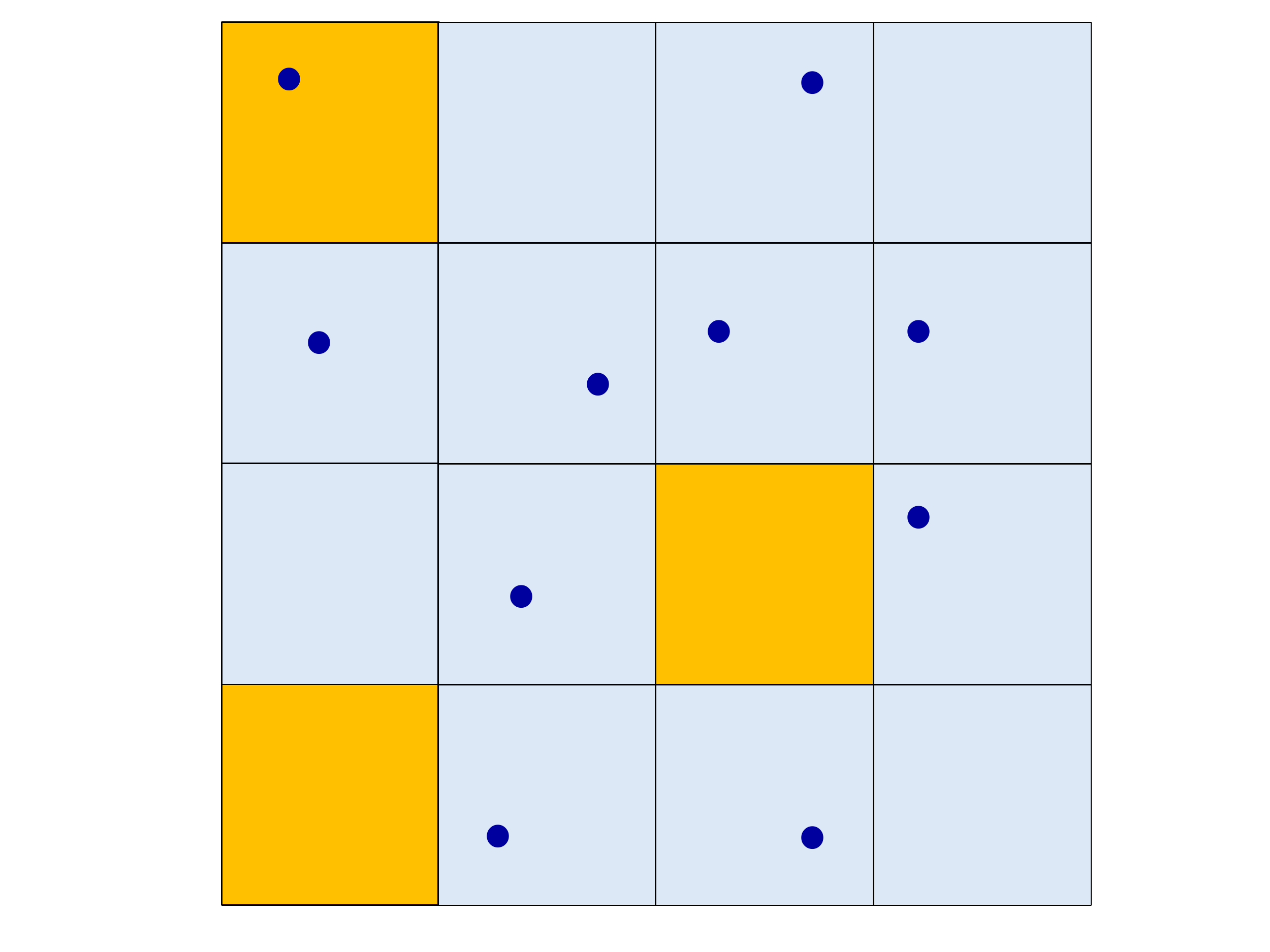}
\centerline{{\small (a) Limitations for small data}}
\end{minipage}
\hfill
\begin{minipage}[b]{0.49\linewidth}
\centering
\includegraphics*[width=4cm,height=3cm]{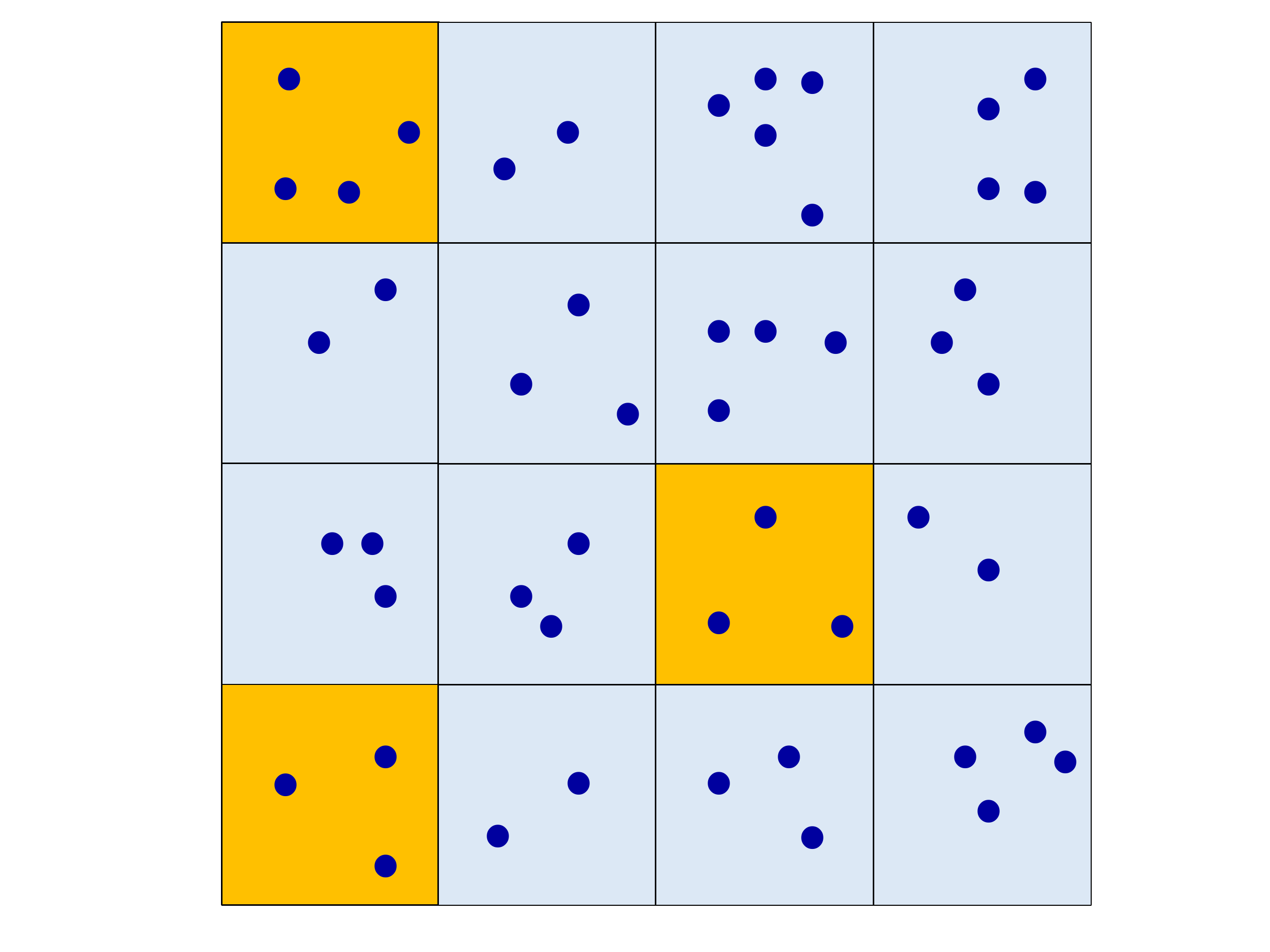}
\centerline{{\small (b) Advantages for massive data}}
\end{minipage}
\hfill
\caption{The role of data size in realizing spatial sparsity}
\label{Figure:sparse-and-data}
\end{figure}

Our purpose is   not only to pursue the power of depth in deep learning, but also to show the important role of the data size  in embodying  advantages of deep nets. To this end, we aim at finding  data feature (or function) that is difficult to be reflected by  data of small size,  but is easily captured by massive data. The spatially sparse feature (or function) naturally comes into our sights. As demonstrated in Figure \ref{Figure:sparse-and-data}, if a function is supported on the orange range, then small data content as shown in Figure \ref{Figure:sparse-and-data} (a) cannot capture the spareness of the support. It requires at least one sample point in each sub-cube to reflect the spatial spareness as shown in Figure 1(b).  Such a spatially sparse assumption abounds in numerous  application regions such as   computer vision \cite{Wright2010}, signal processing \cite{ELayach2014} and pattern recognition \cite{Hou2012}, and several special deep nets have been designed to extract  spatially sparse features of data \cite{Graham2014}.

Due to the   limitation  of small size data-sets in  reflecting the spatial sparseness as shown in Figure \ref{Figure:sparse-and-data}, this paper is devoted to deriving    the quantitative  requirement of the data size to extract the spatial sparseness.   In particular,  we
prove  existence of some learning scheme that can reflect both the smoothness and spatial sparseness, provided that the data size achieves a certain level. This finding coincides with the well-known sampling theorem in compressed sensing \cite{Donoho2006}. We then reformulate our sampling theorem in the framework of learning theory \cite{Cucker2007} by highlighting the important role of data size in deriving optimal learning rates for learning smooth and spatially sparse functions.
The established  sampling theorem  in learning theory theoretically verifies the necessity of massive data in sparseness-related applications and shows that massive data can extract some data features  that cannot   be reflected by data of small size.

By applying the piecewise linear and continuous property of the rectifier linear unit (ReLU) function, $\sigma(t):=\max\{0,t\}$, we construct a deep net with two hidden layers and finitely many neurons to provide a localized approximation, which is beyond the capability of shallow nets \cite{Chui1994,Safran2016,Chui2018}. The localized approximation of deep nets highlights their
  power in capturing the position information of data inputs.
  A direct
consequence is that deep nets can reflect the spatially sparse functions \cite{Lin2018}. This property, together with the recently developed approaches  in approximating smooth function by deep nets \cite{Yarotsky2017,Petersen2017,Han2019}, give rise to the feasibility of adopting deep nets to  extracting smoothness and spatial sparseness simultaneously. We succeed in deriving almost optimal learning rates for implementing empirical risk minimization (ERM) on deep nets and proving that up to a logarithmic factor,  the derived learning rates coincide with those of the sampling theorem. In other words, our results theoretically verify the benefits of massiveness of data in learning smooth and spatially sparse functions, and that deep learning is capable of embodying  advantages of massive data.

 The rest of this paper is organized as
follows.   In Section \ref{Sec.Sampling}, we show the popularity of spatially sparse functions and  present the sampling theorem for learning smooth and spatially sparse functions. In Section \ref{Sec.Deep-nets}, we provide the advantage of deep nets in embodying the benefits of massive data via showing the optimal learning rates for ERM on deep nets. In Section \ref{Sec.Upperbound}, we establish  upper bounds of the sampling theorem and learning rates for ERM on deep nets. In Section \ref{Sec.Lower bound}, we present the proofs for the  lower bounds.

\section{Sampling Theorem for Realizing Spatially Sparse and Smooth Features}\label{Sec.Sampling}
In this section, we discuss the  benefits of massive data via presenting a sampling theorem in the framework of learning theory.

\subsection{Spatially sparse and smooth functions}

\begin{figure}[!t]
\begin{minipage}[b]{0.49\linewidth}
\centering
\includegraphics*[width=3cm,height=3cm]{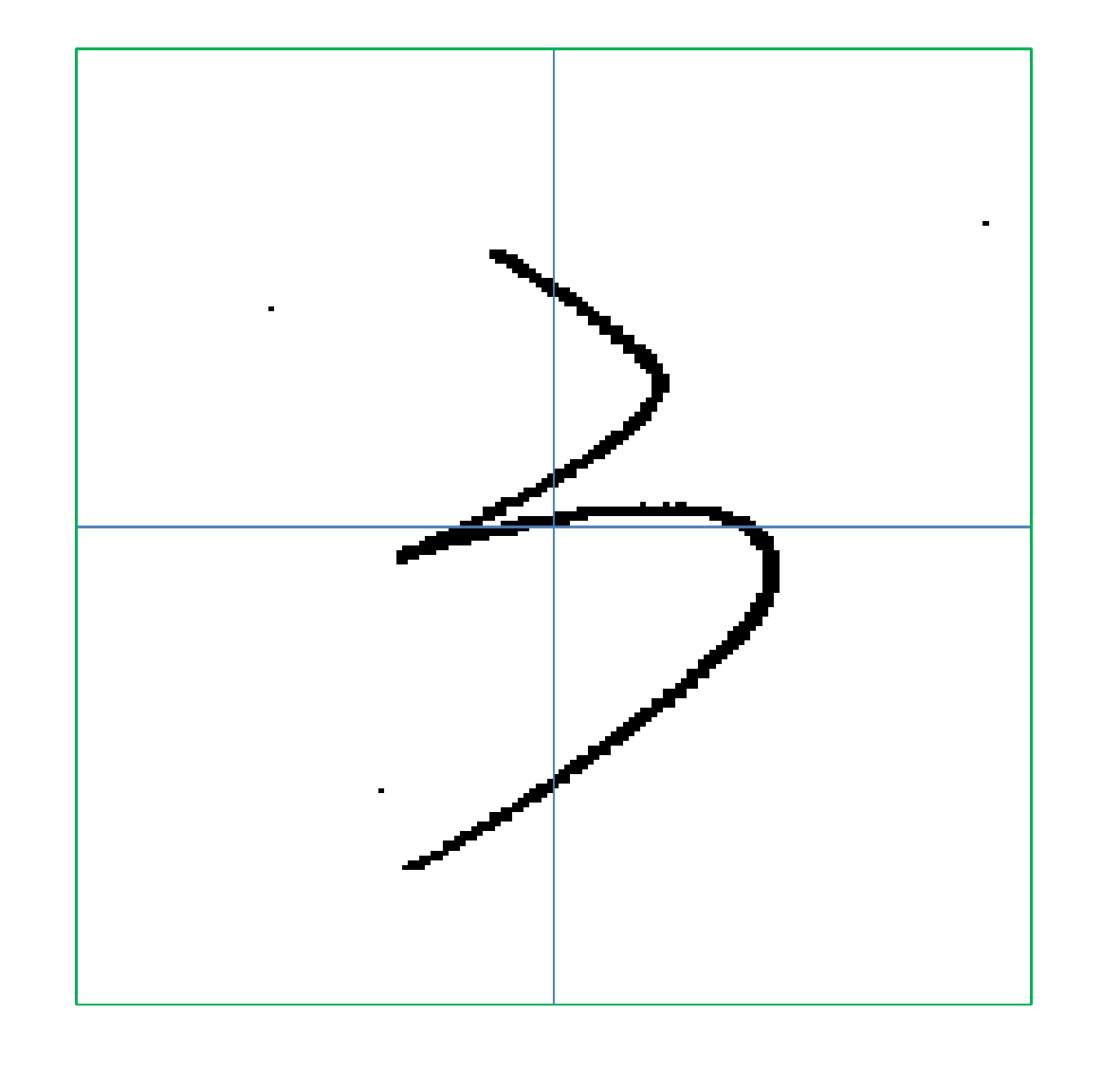}
\centerline{{\small (a) Non-sparseness for partitions}}
\end{minipage}
\hfill
\begin{minipage}[b]{0.49\linewidth}
\centering
\includegraphics*[width=3cm,height=3cm]{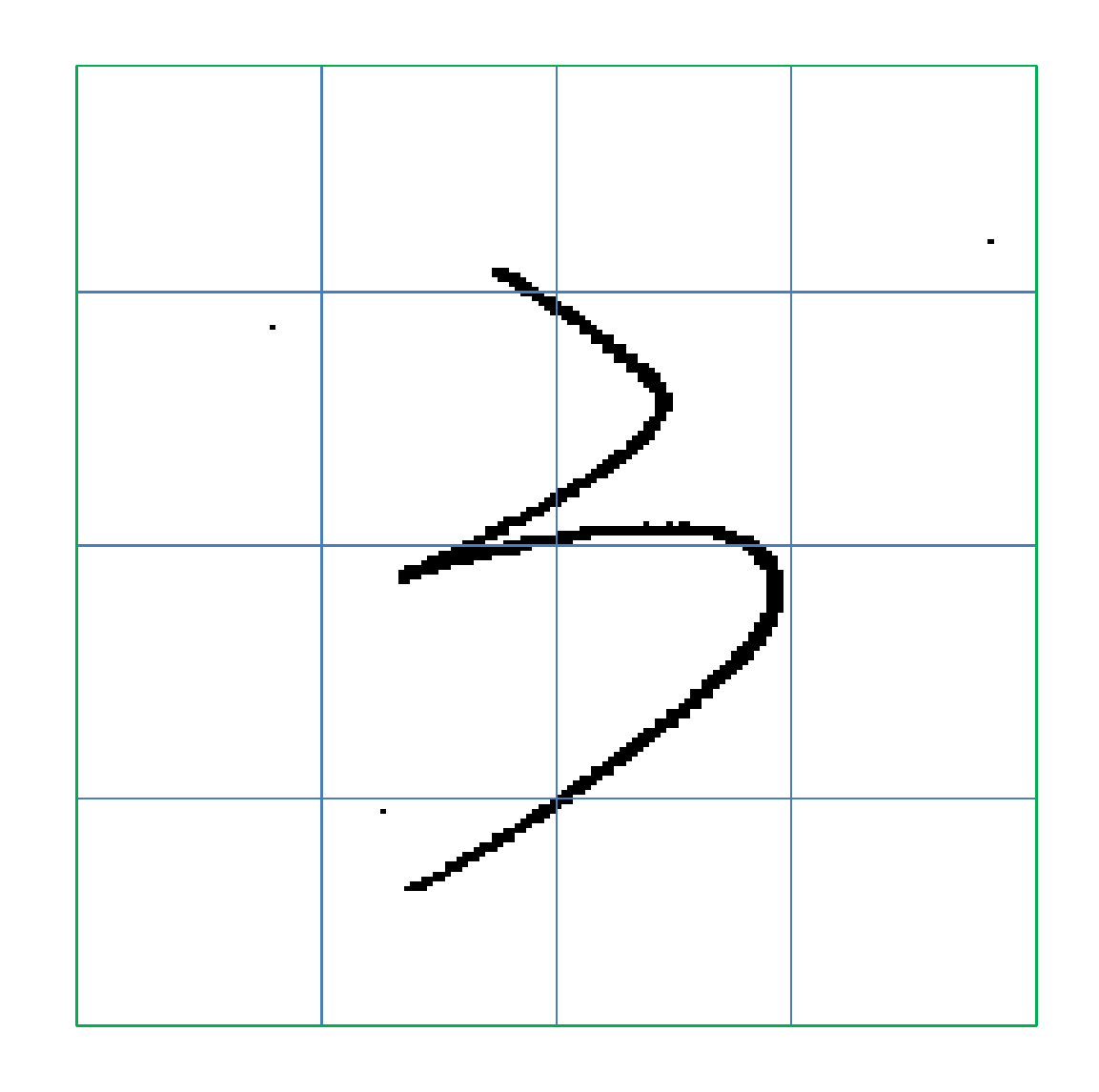}
\centerline{{\small (b) Sparseness for partitions}}
\end{minipage}
\hfill
\caption{The role of partition in reflecting the spatial sparsity}
\label{Figure:sparse-and-partition}
\end{figure}

Spatial  sparseness is a popular data feature  which abounds in numerous applications such as  handwritten digit recognition \cite{Ciresan2010},  magnetic resonance imaging (MRI) analysis \cite{Akkus2017}, image classification \cite{Yang2009}  and environmental data processing  \cite{Deluna2005}. Different from other sparseness measurements such as the sparseness in the frequency domain \cite{LinH2017,Schwab2018} and the manifold sparseness \cite{Chui2018},   spatial sparseness depends heavily on  partitions of the input space. Considering handwritten digit recognition as an example, Figure \ref{Figure:sparse-and-partition}  (a)  shows that the handwritten digit is not sparse if the partition level is $4$. However, if the partition level achieves $16$ as shown in Figure \ref{Figure:sparse-and-partition}  (b), the handwritten digit is sparse.

Based on   this observation, we present the following definition of spatially sparse functions (see, \cite{Lin2018}).
Let $\mathbb I^d:=[0,1]^d$ and $N\in\mathbb N$. Partition $\mathbb
I^d$ by $N^d$ sub-cubes $\{A_j\}_{j=1}^{N^d}$ of side length
$N^{-1}$ and with centers $\{\zeta_j\}_{j=1}^{N^d}$. For $s\in\mathbb
N$ with $s\leq N^d$,
$$
    \Lambda_s:=\left\{j_\ell:
      j_\ell\in\{1,2,\dots,N^d\}, 1\leq \ell\leq s\right\},
$$
and consider a function
$f$ defined on $\mathbb I^d$, if the support of $f$ is contained in
$S:=\cup_{j\in \Lambda_s}A_{  j}$ for a subset $\Lambda_s$ of $\{1,2,\dots,N^d\}$ of cardinality at most $s$. We then say that $f$ is
$s$-sparse in $N^d$ partitions. In what follows, we take $\Lambda_s$ to be the smallest subset that satisfies this condition.

Besides the spatial sparseness, we also introduce the smooth property of $f$, which is a widely used a-priori assumption \cite{Yarotsky2017,Petersen2017,Lin2018CA,Chui2018,Zhou2018}.
Let   $c_0>0$ and $r=u+v$ with $u\in\mathbb N_0:=\{0\}\cup\mathbb N$
and $0<v\leq 1$. We say that a   function $f:\mathbb
I^d\rightarrow\mathbb R$ is $(r,c_0)$-smooth if $f$ is $u$-times
differentiable and for any  $\alpha = (\alpha_1, \cdots, \alpha_d) \in
{\mathbb N}^d_0$ with  $\alpha_1+\dots+\alpha_d=u$ and
          $x,x'\in\mathbb I^d$,  its   partial
derivative, denoted by
$$
         f^{(u)}_\alpha (x)=\frac{\partial^uf}{\partial x_1^{\alpha_1}\dots\partial
          x_d^{\alpha_d}}
          (x),
$$
satisfies the Lipschitz condition
\begin{equation}\label{lip}
          \left|f^{(u)}_\alpha (x)-f^{(u)}_\alpha
          (x')\right|\leq c_0\|x-x'\|^v,
\end{equation}
where  $\|x\|$ denotes the Euclidean norm of
  $x$.
Denote by $Lip^{(r,c_0)}$ the family of $(r,c_0)$-smooth
functions satisfying (\ref{lip}) and by $Lip^{(N,s,r,c_0)}$
the set of all    $f\in Lip^{(r,c_0)}$ which are $s$-sparse in $N^d$
partitions.

\subsection{Sampling theorem for realizing spatially sparse and smooth features}
We conduct the analysis in
a standard least-square regression framework
 \cite{Cucker2007},
in which samples $D=\{(x_i,y_i)\}_{i=1}^m$ are drawn independently
 according to  an unknown Borel probability measure $\rho$ on ${\mathcal Z}
 ={\mathcal X}\times {\mathcal Y}$ with $\mathcal
X=\mathbb I^d$ and $\mathcal Y\subseteq[-M,M]$ for some $M>0$. The
  objective is the regression function defined by
$$
         f_\rho(x)=\int_{\mathcal Y} y d\rho(y|x), \qquad x\in\mathcal
         X,
$$
which minimizes the generalization error
$$
         \mathcal E(f):=\int_{\mathcal Z}(f(x)-y)^2d\rho,
$$
where $\rho(y|x)$ denotes the conditional distribution at $x$
induced by $\rho$.  Let $\rho_X$ be the marginal distribution of
$\rho$ on $\mathcal X$ and $(L^2_{\rho_{_X}}, \|\cdot\|_\rho)$ denote
the Hilbert space of $\rho_X$ square-integrable functions on
$\mathcal X$. Then   for   $f\in L_{\rho_X}^2$, it follows, in view of
\cite{Cucker2007}, that
\begin{equation}\label{equality}
        \mathcal E(f)-\mathcal E(f_\rho)=\|f-f_\rho\|_\rho^2.
\end{equation}

If $f_\rho$ is supported on $S$ but
$\rho_X$ is supported on $\mathbb I^d\backslash S$, it is impossible
to derive a satisfactory learning rate, implying  the  necessity of restrictions  on $\rho_X$. In this section, we assume $\rho_X$ is the uniform distribution for the sake of brevity. Our result also holds   under the classical distortion assumption  on $\rho_X$ \cite{Zhou2006}.
Denote by $\mathcal M(N,s,r,c_0)$  the set of all distributions
satisfying that $\rho_X$ is the uniform distribution and $f_\rho\in Lip^{(N,s,r,c_0)}$.
We enter into a competition over all estimators $\Psi_D:D\rightarrow f_{D}$ and define
\begin{eqnarray*}
          e(N,s,r,c_0)
           := \sup_{\rho\in \mathcal M(N,s,r,c_0)}\inf_{f_D\in \Psi_D}\mathbf E(\|f_\rho-f_{D}\|^2_\rho).
\end{eqnarray*}
The following theorem is our first main result.

\begin{theorem}\label{Theorem:sampling-theorem}
Let  $r,c_0>0$, $d,s,N,m\in\mathbb N$ with $s\leq N^d$.  If
\begin{equation}\label{restions-on-m}
       \frac{m}{\log m}\geq C^*\frac{N^{2r+2d}}{s},
\end{equation}
then
\begin{eqnarray}\label{samplingtheorem}
      &&C_1 m^{-\frac{2r}{2r+d}}\left(\frac{s}{N^d}\right)^\frac{d}{2r+d}\leq
       e(N,s,r,c_0) \nonumber\\
       &\leq&
     C_2
     \left(\frac{m}{\log m}\right)^{-\frac{2r}{2r+d}}\left(\frac{s}{N^d}\right)^\frac{d}{2r+d},
\end{eqnarray}
where $C^*$, $C_1$, $C_2$ are constants
 independent of $m$, $s$ or $N$.
\end{theorem}

The proof of Theorem \ref{Theorem:sampling-theorem} will be given in Sec. \ref{Sec.Lower bound}.
The sampling theorem \cite{Shannon1949} originally focuses on deriving   the minimal sampling rate   that permits a discrete sequence of samples to capture all the information from a continuous-time signal of finite bandwidth in sampling processes. Recent developments \cite{Zayed2018} imitate the sampling theorem in terms of deriving minimal sizes of samples to represent a signal via some transformations such as  wavelet, Fourier and Legendre transformations. In learning theory, the sampling theorem studied in this paper aims at deriving minimal sizes of samples that can achieve the optimal learning rates for some specified learning task. Theorem \ref{Theorem:sampling-theorem} shows that  optimal learning rates for learning spatially sparse and smooth functions are achievable provided (\ref{restions-on-m}) holds.  The size of samples, as governed in (\ref{restions-on-m}), depends on the sparsity level $s$ and partitions numbers $N$, and increases with respect to $N$, showing that more partitions require more samples. This coincides with the intuitive observation as shown in Figure \ref{Figure:sparse-and-data}. Different from the classical results in signal processing \cite{Zayed2018}, the size of  samples in (\ref{restions-on-m}) decreases  with   $s$. This is not surprising, since   the established  optimal learning rates in (\ref{samplingtheorem})
increase  with $s$. In  other words, the size of samples in our result is to recognize the support of the regression function and thus increases with $N$ while the sparsity $s$ is reflected by   optimal learning rates in (\ref{samplingtheorem}).

The almost optimal learning rate in (\ref{samplingtheorem}) can be regarded as a combination of  two components $m^{-\frac{2r}{2r+d}}$ for the smoothness and $\left(\frac{s}{N^d}\right)^\frac{d}{2r+d}$ for the sparseness. If $s=N^d$, meaning that $f_\rho$ is not spatially sparse, then the learning rate  derived in Theorem \ref{Theorem:sampling-theorem} coincides with the optimal learning rate in learning smooth functions (\cite[Chap. 3]{Gyorfi2002}), up to a logarithmic factor. If $r$ is extremely small, the   learning rate  derived in Theorem \ref{Theorem:sampling-theorem}, near to $\frac{s}{N^d}$ due to the uniform assumption on $\rho_X$, is also the optimal learning rates for learning spatially sparse functions. If $m$ is relatively small with respect to $N$, i.e. (\ref{restions-on-m}) does not hold, then while the smoothness part $m^{-\frac{2r}{2r+d}}$ can be maintained,  the sparseness property cannot be captured. This shows the benefit of massive data in learning spatially sparse functions. It should be noted that there is an additional logarithmic term in (\ref{samplingtheorem}). We believe that it is removable by using different tools from this paper and will consider it as a future work.

\section{Deep Nets in Realizing Spatial Spareness}\label{Sec.Deep-nets}
In this section, we verify the power of depth for ReLU nets in localized approximation and    spatially sparse approximation, and then show that deep nets are able to embody the benefits of massive data in learning spatially sparse and smooth functions.

\subsection{Deep ReLU nets}
One of the main reasons for the great success of deep learning is the implementation in terms  of deep nets. In comparision  with the classical shallow nets, deep nets are significantly better  in providing localized approximation \cite{Chui1994}, manifold learning \cite{Shaham2015,Chui2018}, realizing rotation invariance priors \cite{McCane2017,Chui2018a}, embodying sparsity in the frequency domain \cite{LinH2017,Schwab2018} and in the spatial domain \cite{Lin2018}, approximating piecewise smooth functions \cite{Petersen2017} and capturing the hierarchical structures \cite{Mhaskar2016a,Kohler2017} etc.. However, all these interesting results are not yet sufficient to explain
  why deep nets perform well in the era of big data.

Let $\sigma(t):=\max\{t,0\}$ be the rectifier liner unit (ReLU).
Deep ReLU nets, i.e. deep nets with the ReLU activation function, is most popular
in current research in deep learning. Due to the non-differentiable property of ReLU, it seems difficult for ReLU nets to approximate smooth functions at the first glance. However, it was shown in  \cite{Yarotsky2017,Petersen2017,Zhou2018a,Han2019}   that increasing the depth of ReLU nets succeeds in overcoming this problem and thus provides theoretical foundations in understanding deep ReLU nets.

Denote
$x=(x^{(1)},\dots,x^{(d)})\in\mathbb I^d$. Let
$L\in\mathbb N$ and $d_0,d_1,\dots,d_L\in\mathbb N$ with $d_0=d$.
For
$\vec{h}=(h^{(1)},\dots,h^{(d_k)})^T\in\mathbb R^{d_k}$, define
$\vec{\sigma}(\vec{h})=(\sigma(h^{(1)}),\dots,\sigma(h^{(d_k)}))^T$.
Deep ReLU nets with depth $L$ and width $d_j$ in the $j$-th hidden layer can be mathematically
represented as
\begin{equation}\label{Def:DFCN}
     h_{\{d_0,\dots,d_L,\sigma\}}(x)=\vec{a}\cdot
     \vec{h}_L(x),
\end{equation}
where
\begin{equation}\label{Def:layer vector}
    \vec{h}_k(x)=\vec{\sigma}(W_k\cdot
    \vec{h}_{k-1}(x)+\vec{b}_k),\qquad k=1,2,\dots,L,
\end{equation}
$\vec{h}_{0}(x)=x,$ $\vec{a}\in\mathbb R^{d_L}$,
$\vec{b}_k\in\mathbb R^{d_k},$
 and $W_k=(W_k^{i,j})_{i=1,j=1}^{d_{k},d_{k-1}}$
is a $d_{k}\times
 d_{k-1}$ matrix. Denote by $\mathcal H_{\{d_0,\dots,d_L,\sigma\}}$ the set of
all these deep ReLU nets. The structures of deep nets are reflected by  weight matrices $W_k$ and threshold  vectors $\vec{b}_k$, $k=1,\dots,L$. For example,  taking the special form of  Toeplitz-type weight matrices   leads to the deep convolutional nets
\cite{Zhou2018,Zhou2018a,Zhou2019}, full matrices correspond to deep fully connected nets \cite{Goodfellow2016}, and tree-type sparse matrices imply deep nets with tree structures \cite{Chui2018a,Chui2019}. In this paper, we do not focus on the structure selection of deep nets, but rather on the existence of some deep net structure for realization of  the sampling theorem established in Theorem \ref{Theorem:sampling-theorem}.

\subsection{Deep ReLU nets for localized approximation  }

Localized approximation is an important  property  of neural networks in that  it is a crucial  step-stone in   approximating
piecewise smooth functions \cite{Petersen2017} and spatially sparse functions
  \cite{Lin2018}.
The localized approximation of a neural network  allows the
target function to be modified in any small region of the Euclidean
space by adjusting a few neurons, rather than the entire network.
It was originally proposed in \cite[Def. 2.1]{Chui1994} to demonstrate the power of depth for deep nets with  sigmoid-type activation functions. The main conclusion in \cite{Chui1994} is that deep nets only with two hidden layers and $2d+1$ neurons can provide localized approximation, while shallow nets fail for $d\geq 2$, even for the most simple Heaviside activation function. In this section,   we prove that deep ReLU nets with two hidden layers and $4d+1$ neurons are capable of providing  localized approximation.

 For
 $a,b\in\mathbb R$ with $ a<b $, define a trapezoid-shaped
function $T_{\tau,a,b}$ with a parameter $0<\tau\leq 1$ as
\begin{eqnarray}\label{trapezoid function}
   &&T_{\tau,a,b}(t):=\frac1\tau\big\{\sigma(t-a+\tau)-\sigma(t-a) \nonumber\\
   &-&
   \sigma(t-b)+\sigma(t-b-\tau)\big\}.
\end{eqnarray}
Then the definition of $\sigma$ yields
\begin{equation}\label{Deteailed trapezoid}
   T_{\tau,a,b}(t)=\left\{\begin{array}{cc}
   1,&\mbox{if}\ a\leq t\leq b,\\
   0,& \mbox{if}\ t\geq b+\tau,\ \mbox{or}\ t\leq a-\tau,\\
   \frac{b+\tau-t}{\tau}, &\mbox{if}\ b<t<b+\tau,\\
   \frac{t-a+\tau}{\tau}, & \mbox{if}\ a-\tau<t<a.
   \end{array}
   \right.
\end{equation}
We may then consider
\begin{eqnarray}\label{Def.N1}
    N_{a,b,\tau}(x) := \sigma\left(\sum_{j=1}^dT_{\tau,a,b}(x^{(j)})-(d-1)\right).
\end{eqnarray}
The following proposition presents the localized approximation
property of $N_{a,b,\tau}$.

\begin{proposition}\label{Proposition:Local approximation}
Let  $ a<b$, $0<\tau\leq 1$ and $N_{a,b,\tau}$ be defined by (\ref{Def.N1}).
Then we have $0\leq N_{a,b,\tau}(x)\leq 1$ for all $x\in\mathbb I^d$
and
\begin{equation}\label{Localized approximation}
     N_{a,b,\tau}(x)=\left\{\begin{array}{cc}
     0,&\mbox{if}\ x\notin[a-\tau,b+\tau]^d,\\
     1,&\mbox{if}\ x\in [a,b]^d.
     \end{array}
     \right.
\end{equation}
\end{proposition}

\begin{figure}[!t]
\begin{minipage}[b]{0.49\linewidth}
\centering
\includegraphics*[width=4cm,height=3cm]{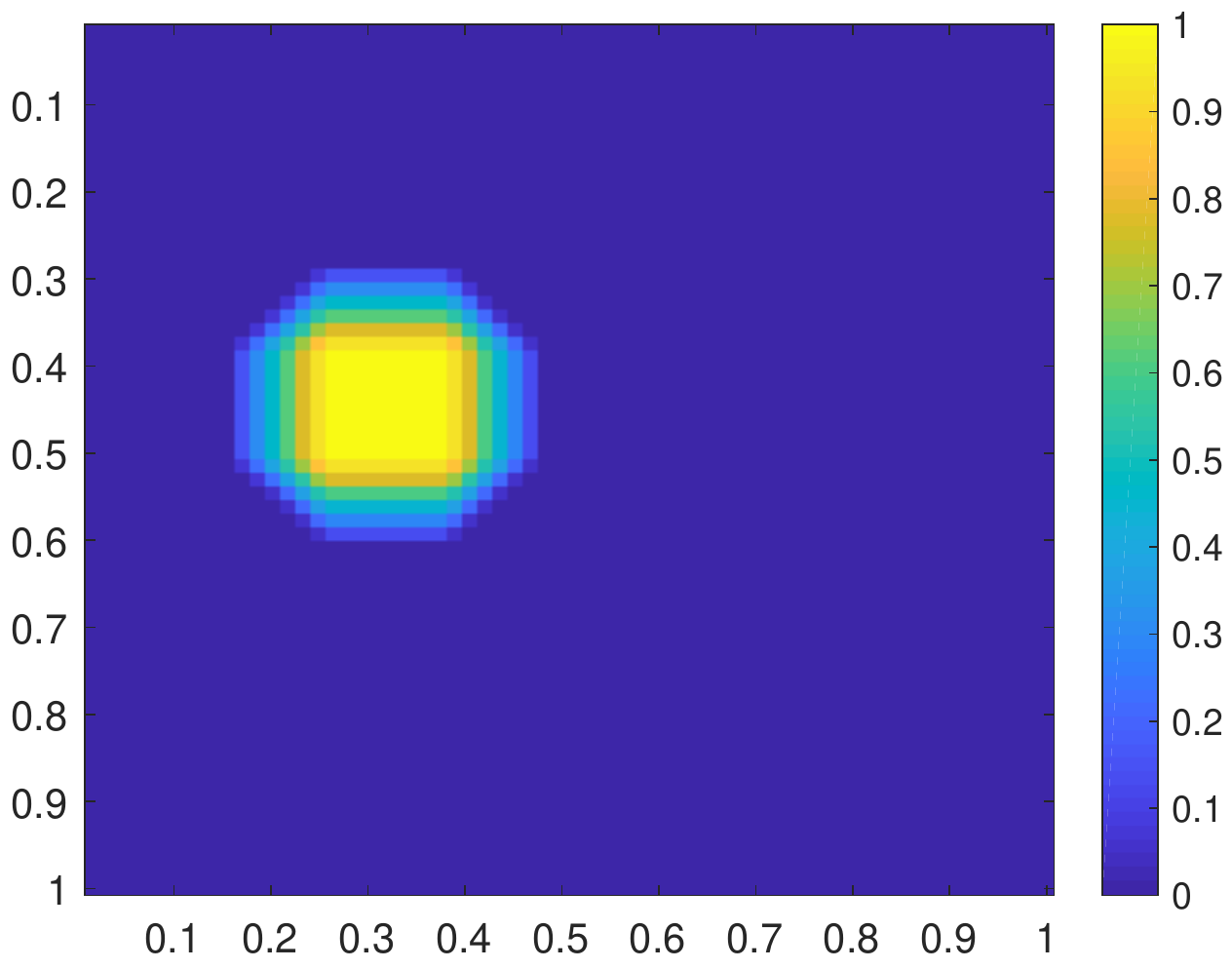}
\centerline{{\small (a)  $\tau=0.1$}}
\end{minipage}
\hfill
\begin{minipage}[b]{0.49\linewidth}
\centering
\includegraphics*[width=4cm,height=3cm]{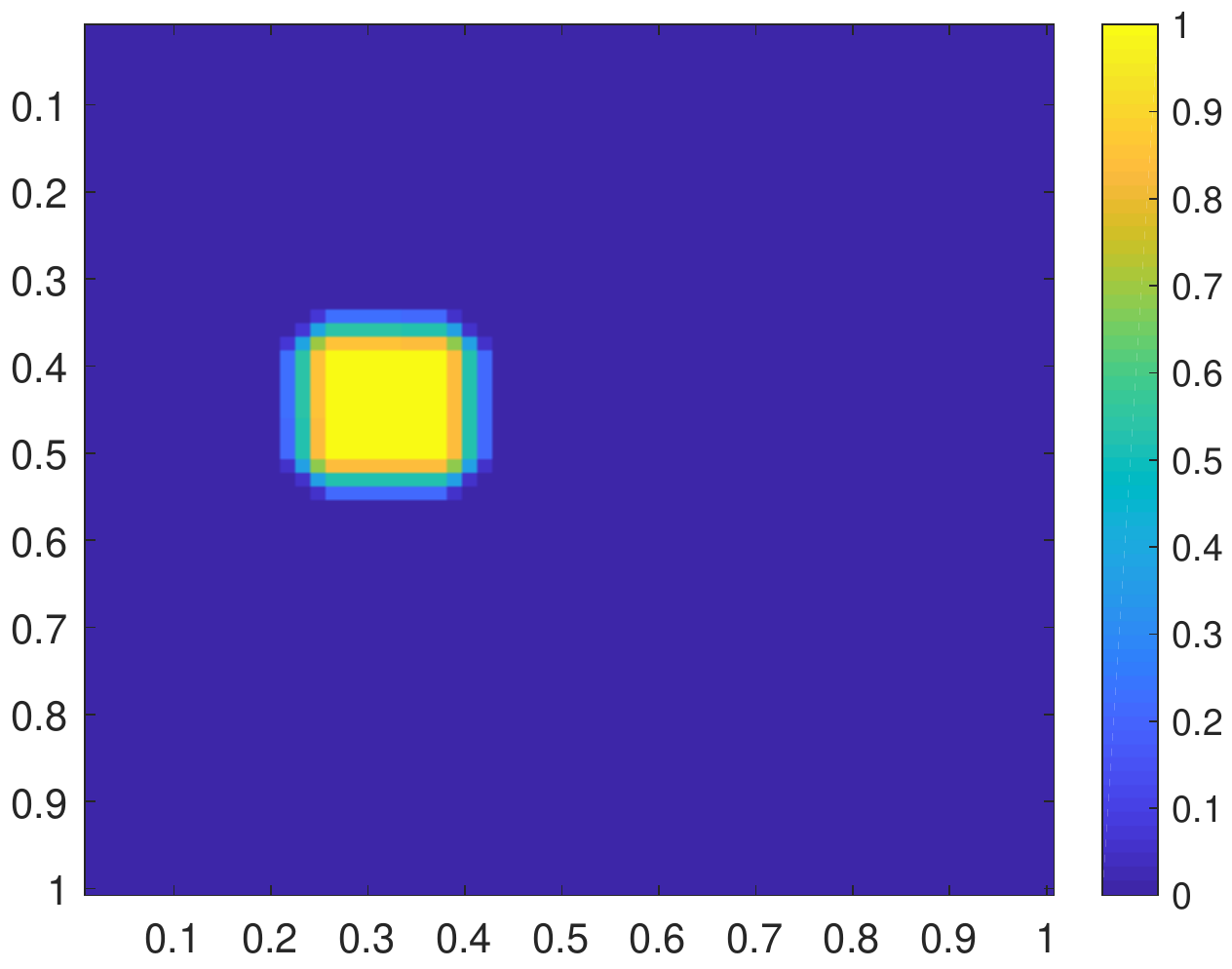}
\centerline{{\small (b)  $\tau=0.05$}}
\end{minipage}
\hfill
\begin{minipage}[b]{0.49\linewidth}
\centering
\includegraphics*[width=4cm,height=3cm]{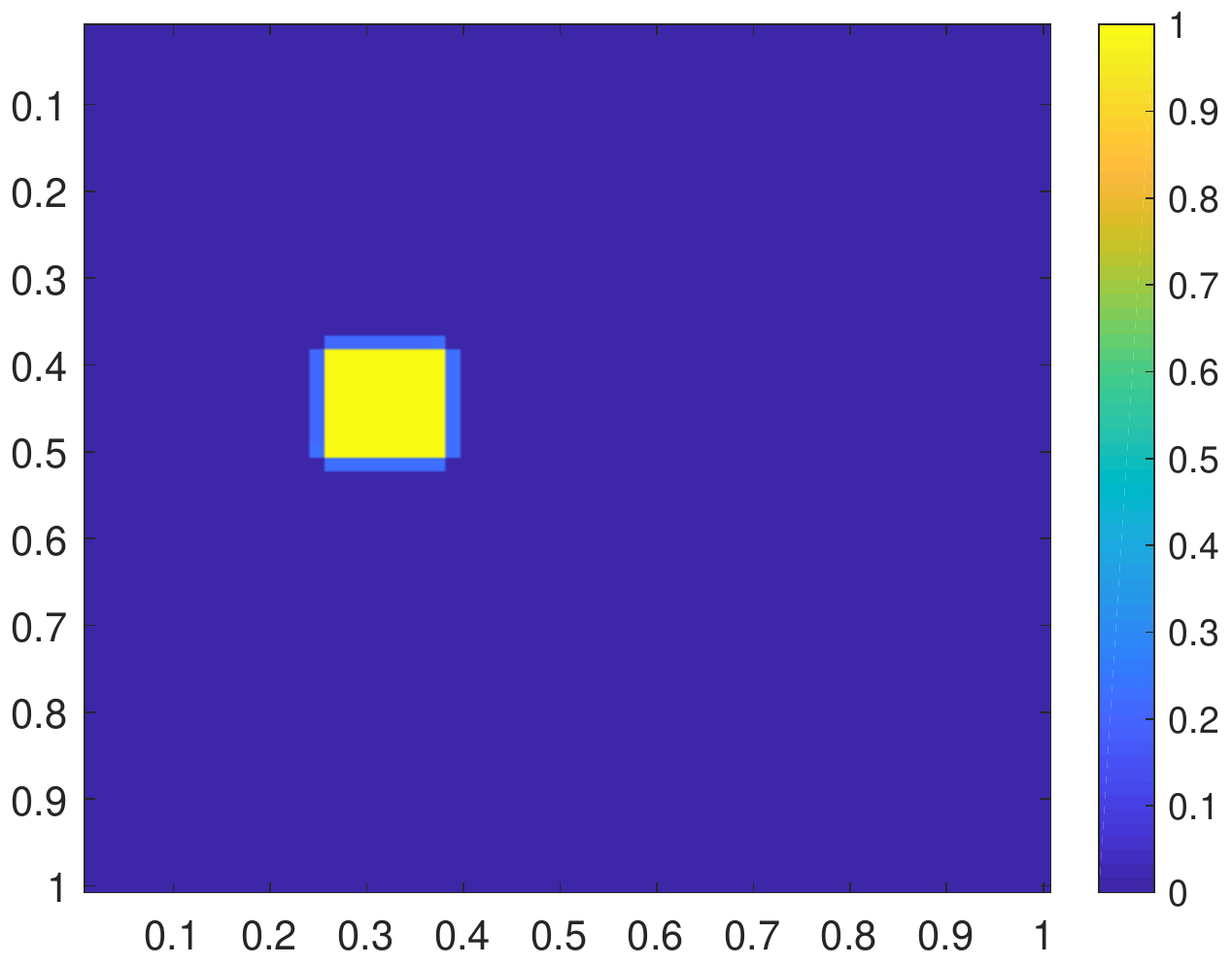}
\centerline{{\small (c)  $\tau=0.01$}}
\end{minipage}
\hfill
\begin{minipage}[b]{0.49\linewidth}
\centering
\includegraphics*[width=4cm,height=3cm]{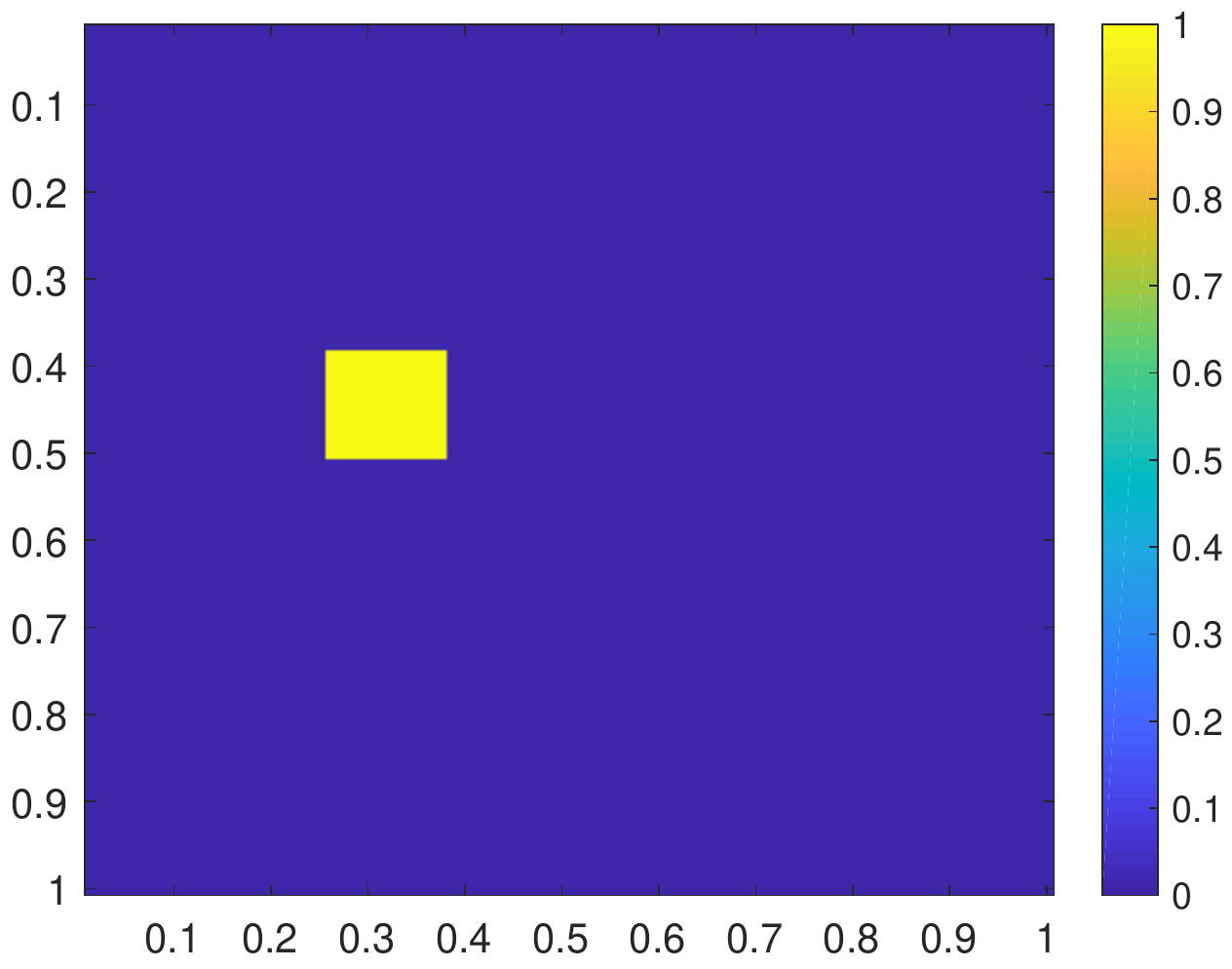}
\centerline{{\small   (d)   $\tau=0.005$}}
\end{minipage}
\hfill
\caption{The localized approximation   based on a cubic partition of $[0,1]^2$ with side length
$1/8$  for the deep net constructed in (\ref{NN-for-localization}) with $\xi=(3/16,5/16)$}
\label{Figure:localized}
\end{figure}

The proof of Proposition \ref{Proposition:Local approximation} will be postponed to Section \ref{Sec.Upperbound}. Similar approximation results for   deep nets with sigmoid-type activation functions and $2d+1$ neurons have been established in \cite{Chui1994,Shaham2015,Lin2018}. The representation in Proposition \ref{Proposition:Local approximation} is better because the expression for $x\in[a,b]^d$ and $x\notin [a-\tau,b+\tau]^d$ is exact.
For arbitrary $N^*\in\mathbb N$, partition $\mathbb I^d$ into
${(N^*)^d}$ sub-cubes $\{B_k\}_{k=1}^{(N^*)^d}$ of side length $
1/N^*$ and with centers $\{\xi_k\}_{k=1}^{(N^*)^d}$.
Write
$\tilde{B}_{k,\tau}:=[\xi_k+[-1/(2N^*)-\tau,1/(2N^*)+\tau]]^d\cap\mathbb I^d$.
  It is obvious
that $B_k\subset \tilde{B}_{k,\tau}$. Define $N_{1, N^*,
\xi,\tau}:  \mathbb I^d \to \mathbb R$ for $\xi\in\mathbb I^d$ by
\begin{eqnarray}\label{NN-for-localization}
      N_{1,N^*, \xi,\tau}(x)=N_{-1/(2N^*),1/(2N^*),\tau}(x-\xi).
\end{eqnarray}
In view of   Proposition \ref{Proposition:Local
approximation}, (\ref{Deteailed trapezoid}) and (\ref{NN-for-localization}), we have
$|N_{1,N^*,\xi_{k},\tau}(x)|\leq 1$ for all $x\in\mathbb I^d$,
$k\in\{1,\dots,(N^*)^d\}$ and
\begin{equation}\label{property localization}
     N_{1,N^*, \xi_{k},\tau}(x)=\left\{\begin{array}{cc}
     0,&\mbox{if}\ x\notin \tilde{B}_{k,\tau},\\
     1,&\mbox{if}\ x\in B_k.
     \end{array}
     \right.
\end{equation}
As shown in
Figure \ref{Figure:localized}, the parameter $\tau$ determines the size of $\tilde{B}_{k,\tau}$, and thus affects the performance of localized approximation for the constructed deep nets in (\ref{NN-for-localization}). However, it does not mean the smaller $\tau$ the better, since the norms of weights decrease with respect to $\tau$, which may result in extremely large capacity of deep ReLU nets for too small $\tau$.

\subsection{Deep ReLU nets for spatially sparse approximation}

\begin{figure}[!t]
\begin{minipage}[b]{0.49\linewidth}
\centering
\includegraphics*[width=4cm,height=3cm]{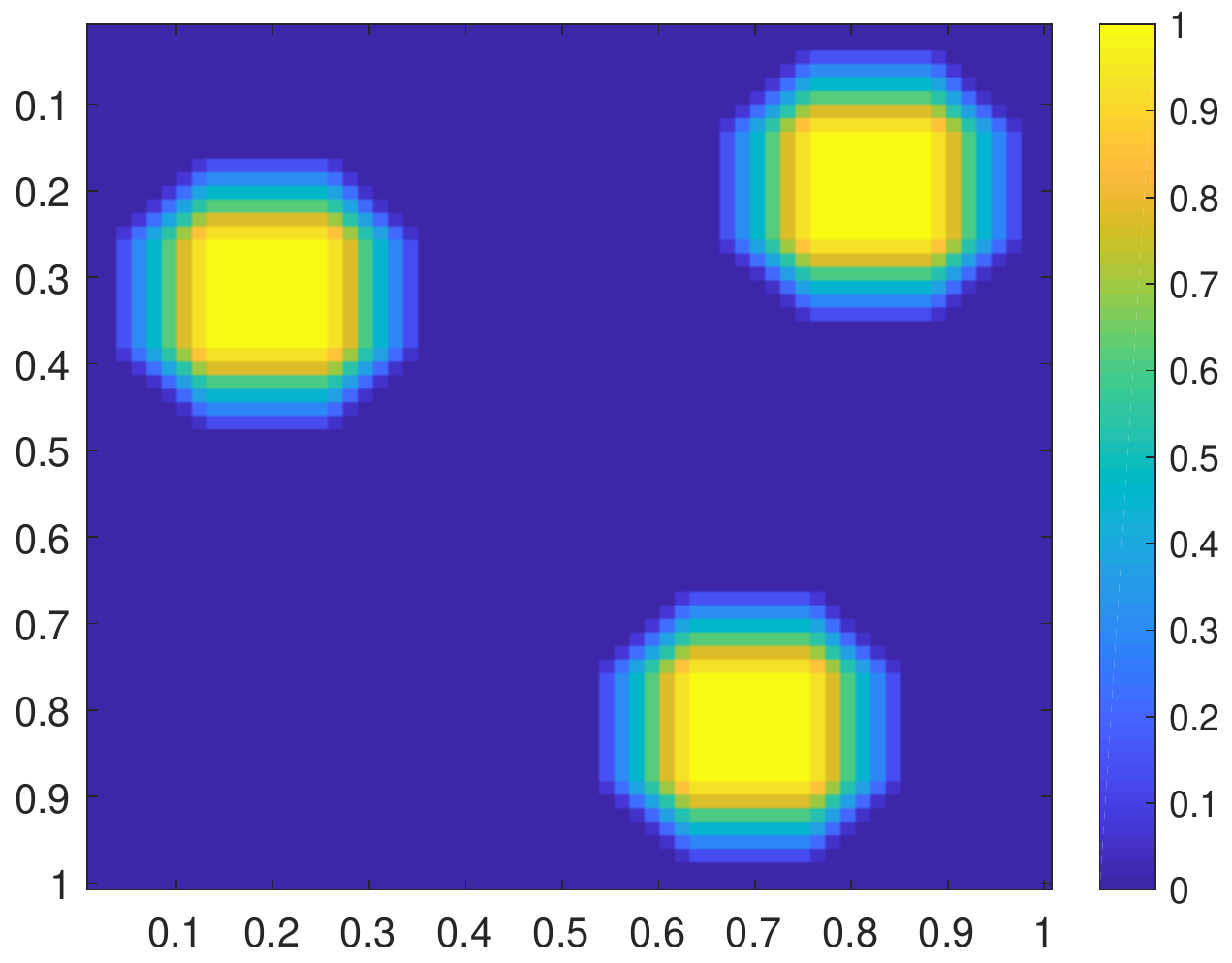}
\centerline{{\small (a)  $\tau=0.1$}}
\end{minipage}
\hfill
\begin{minipage}[b]{0.49\linewidth}
\centering
\includegraphics*[width=4cm,height=3cm]{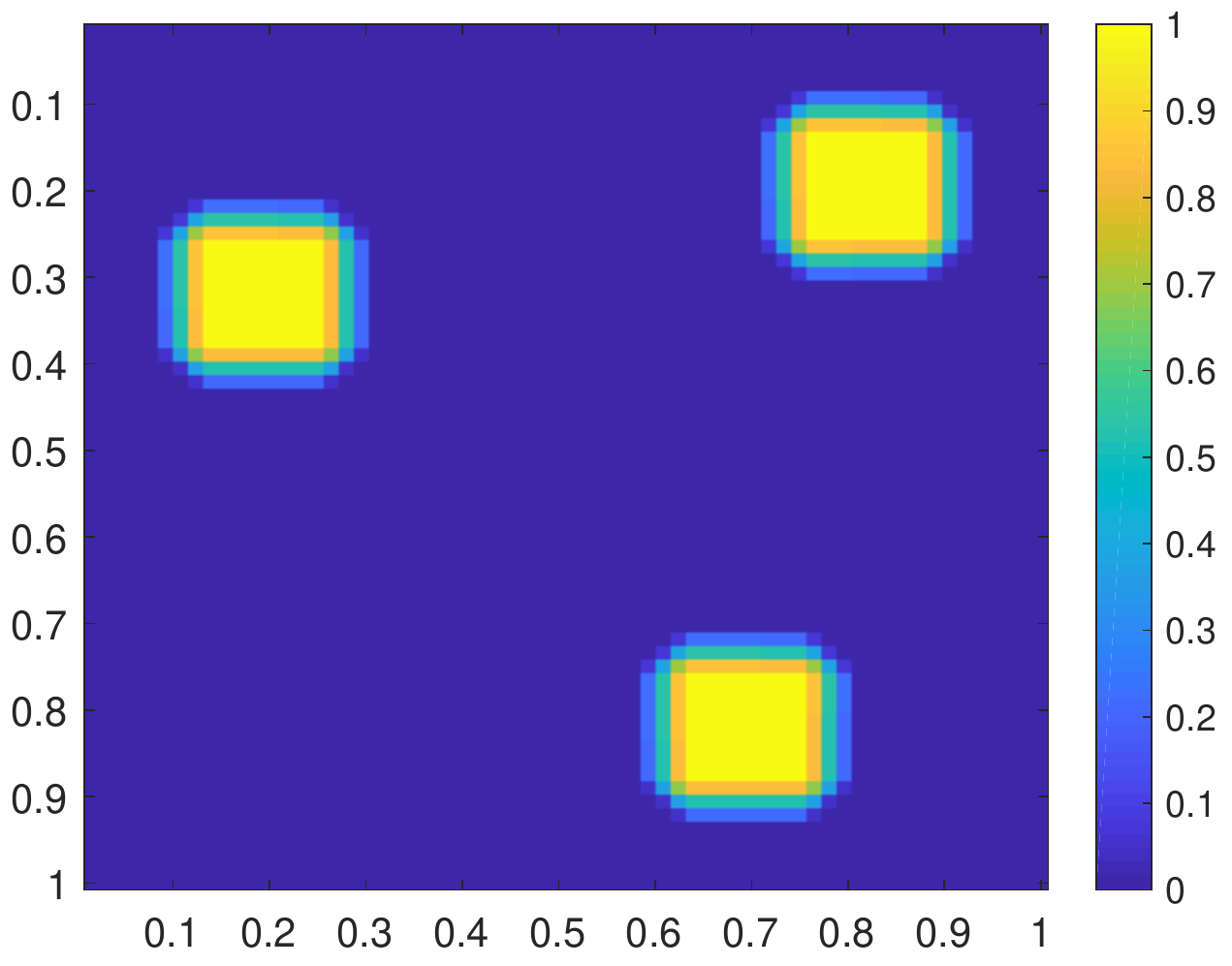}
\centerline{{\small (b)  $\tau=0.05$}}
\end{minipage}
\hfill
\begin{minipage}[b]{0.49\linewidth}
\centering
\includegraphics*[width=4cm,height=3cm]{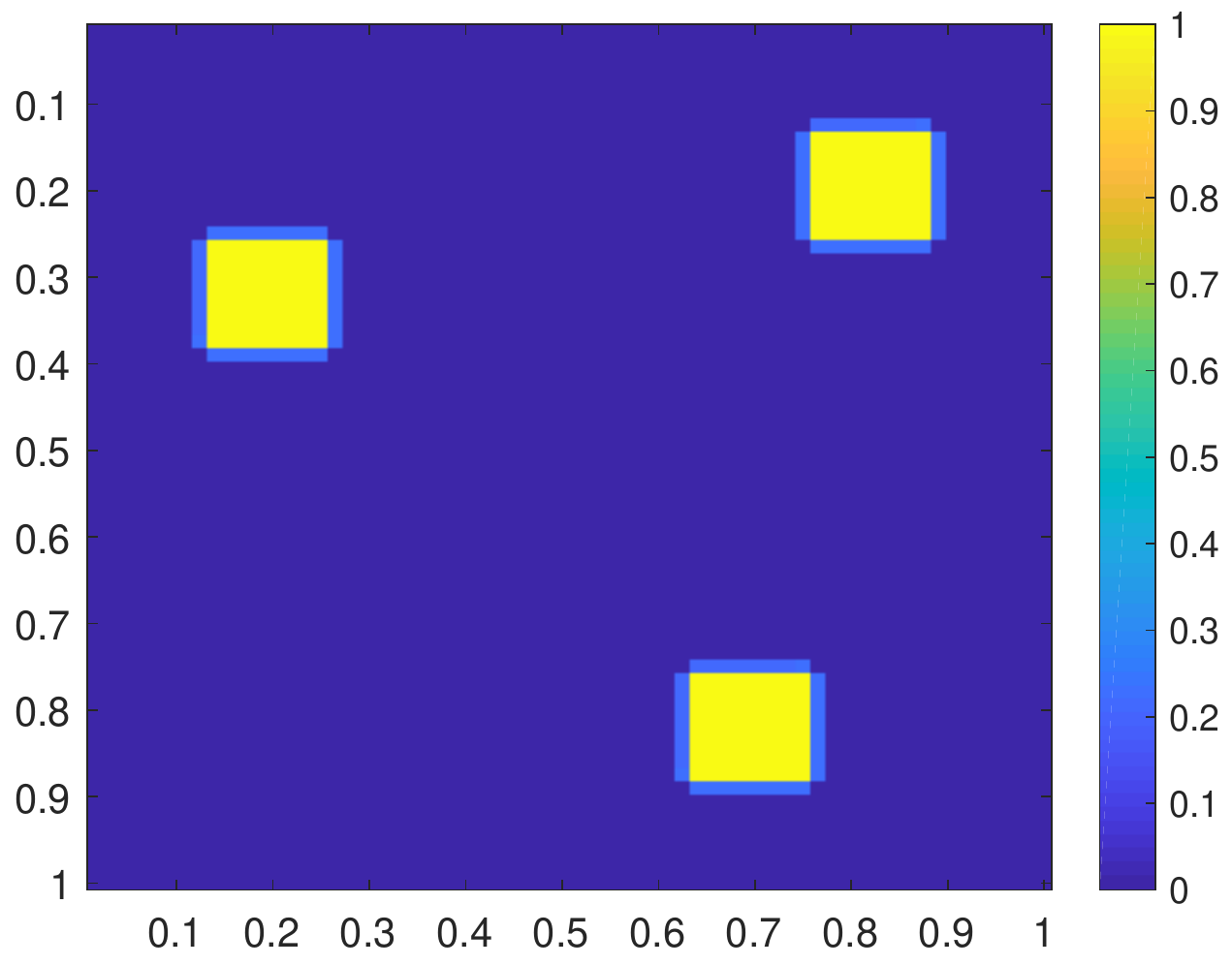}
\centerline{{\small (c)  $\tau=0.01$}}
\end{minipage}
\hfill
\begin{minipage}[b]{0.49\linewidth}
\centering
\includegraphics*[width=4cm,height=3cm]{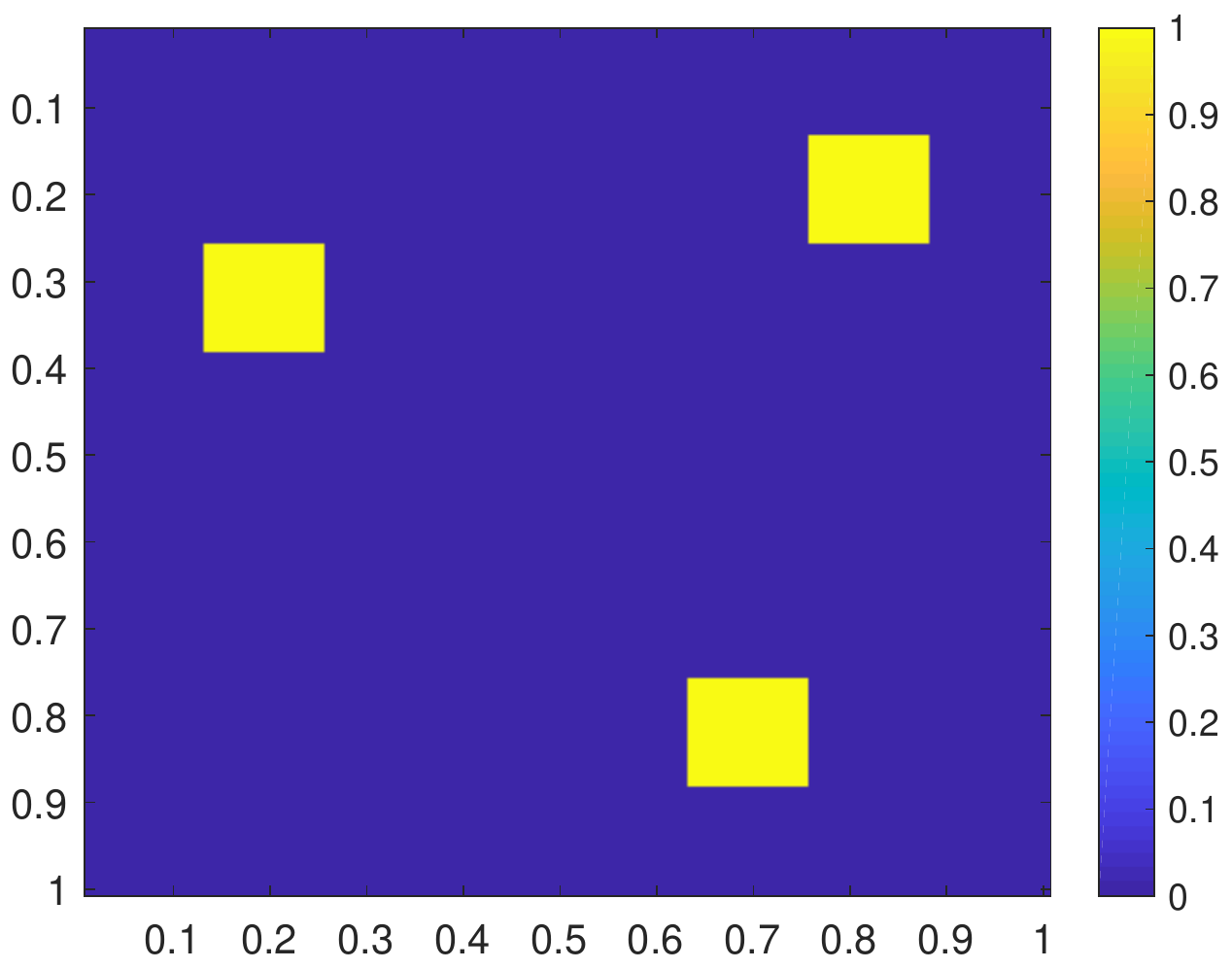}
\centerline{{\small   (d)   $\tau=0.005$}}
\end{minipage}
\hfill
\caption{Realizing spatial  sparseness by  summations of the deep net constructed in (\ref{NN-for-localization}) with different $k$}
\label{Figure:sparse}
\end{figure}

The localized approximation  established in Proposition \ref{Proposition:Local approximation}   demonstrates the
power of deep ReLU nets with two hidden layers  to recognize some spatial information of the input.
A direct consequence is that deep ReLU nets succeed in capturing  the spatially sparse property of functions and also maintaining the capability of deep ReLU nets in approximating smooth functions. On one hand, spatial sparseness  defined in this paper is built upon a cubic partition of $\mathbb I^d$, i.e. $\mathbb I^d=\cup_{j=1}^{N^d}A_j$. If  $N^*\geq  N$, then $A_j\subseteq \cup_{k:A_j\cap B_k\neq\varnothing}$ can be recognized by the localized approximation of $N_{1,N^*, \xi_{k},\tau}$. Figure \ref{Figure:sparse} demonstrates that for small enough $\tau$,  summations of
$N_{1,N^*, \xi_{k},\tau}$ with different $k$ can reflect the spatial   sparseness for $N^*=N$. On the other hand, due to the localized approximation of $N_{1,N^*, \xi_{k},\tau} (x)$, for any $x\in\mathbb I^d$, there is at most $2^d$ indices  $k_j$ with $N_{1,N^*, \xi_{k_j},\tau}(x)=1$ for $j=1,2,\dots,2^d$ and $|N_{1,N^*, \xi_{k},\tau}(x)|$  extremely small for $k\neq k_j$. Then, for large enough $N^*$, the smoothness of $f$ leads to small approximation error for $\left|f(x)-\sum_{k=1}^{(N^*)^d}f(\xi_k)N_{1,N^*, \xi_{k},\tau}(x)\right|$.

With the above observations, we find that deep ReLU nets are capable of realizing both the smoothness and spatial sparseness, which is beyond the capability of shallow ReLU nets \cite{Chui1994,Yarotsky2017}. The following proposition is the main result in this subsection.

\begin{proposition}\label{Proposition:Approximation error for large r}
Let $1\leq p<\infty$, $r,c_0>0$, $N,s,d\in\mathbb N$ with $s\leq N^d$ and $N^*\geq \max\{4N,\tilde{C}\}$.  Then there exists a deep ReLU net structure with $\lceil 25+4r/d+2r^2/d+10r\rceil$ inner layers and at most $C_1^*(N^*)^d$ free  parameters,
such that for any $f\in Lip^{(N,s,r,c_0)}$  and any $0<\tau \leq
 \frac{s}{2N^d(N^*)^{1+pr}}$, there is a deep ReLU net $N_{3,N^*,\tau}$ with the aforementioned structure and free parameters bounded by
\begin{eqnarray}\label{boundofweights}
   \tilde{B}^*:= C_3\max\{1/\tau,(N^*)^{(2d+r)\gamma}\}.
\end{eqnarray}
 such that
\begin{equation}\label{app theorem for large r}
   \|f-N_{3,N^*,\tau}\|_{L^p(\mathbb I^d)}
   \leq C_4(N^*)^{-r}\left(\frac{s}{N^d}\right)^{1/p},
\end{equation}
and
\begin{equation}\label{bound-N3-NN}
    \|N_{3,N^*,\tau}\|_{L^\infty(\mathbb I^d)}\leq C_5,
\end{equation}
where  $\gamma$, $\tilde{C},C_1^*,C_3,C_4,C_5$  are  constants depending only on $c_0$, $r$, $d$ and $\|f\|_{L^\infty(\mathbb I^d)}$.
\end{proposition}

The proof of Proposition \ref{Proposition:Approximation error for large r} will be given in Section \ref{Sec.Upperbound}. Approximating   functions in $Lip^{(r,c_0)}$ is a classical topic in neural network  approximation. It is shown in \cite{Mhaskar1996} that for shallow nets with $C^\infty$ sigmoid  type activation functions and $(N^*)^d$ free parameters, an approximation rate of order $(N^*)^{-r}$ can be achieved. Furthermore, \cite{Maiorov1999b,Lin2017a} provide a lower bound. Although these nice results show the excellent approximation capability of shallow nets, the weights of shallow nets in \cite{Mhaskar1996,Maiorov1999b} are extremely large, resulting in extremely large capacity.  With such extremely large weights, it follow from the results in \cite{Maiorov1999b,Ismailov2014} that there exists a deep net with two hidden layers and finitely many neurons possessing the universal approximation property. The extremely large weights problem can be avoided by deepening the neural networks. In fact, it can be found in \cite{Yarotsky2017,Petersen2017,Han2019} that similar results hold for deep ReLU nets with a few hidden layers and controllable weights, i.e.,   weights increasing polynomially fast with respect to the number of free parameters.  Our Proposition \ref{Proposition:Approximation error for large r} also implies this finding by setting $s=(N^*)^d$ and larger value of  $\tau$. It will be shown in the next subsection that  controllable weights play a crucial role in deriving small variance as well as fast learning rates for implementing ERM on deep ReLU nets.

The approximation rates established in (\ref{app theorem for large r}) not only reveal the power of depth in approximating smooth functions, but also exhibit the advantage of deep ReLU nets in embodying the spatial  sparseness by means of multiplying an additional $\left(\frac{s}{N^d}\right)^{1/p}$ on the optimal approximation rates $(N^*)^{-r}$ for smooth functions. Noting that shallow nets with the Heaviside activation function \cite{Chui1994} cannot provide localized approximation, corresponding to a special case of $s=1$, Proposition \ref{Proposition:Approximation error for large r} show the power of depth of deep ReLU net under the condition $N^*\geq 4N$.

 \subsection{Realizing  optimal learning rates in sampling theorem by deep nets}
In this subsection, we aim at developing a learning scheme to take advantage of the  power of deep ReLU nets in realizing the spatial  sparseness and smoothness. Denote by $\mathcal H_{n,L}$ the collection of deep nets that possess the structure in Proposition \ref{Proposition:Approximation error for large r} with
\begin{equation}\label{hypoth1}
    L=\lceil 25+4r/d+2r^2/d+10r\rceil,\qquad\mbox{and}\quad n=C_1^*(N^*)^d.
\end{equation}
Define
\begin{eqnarray}\label{Hypothesis space}
        &&\mathcal H_{n,L,\mathcal R}:=
       \{h_{n,L}
      \in \mathcal H_{n,L}:|w_{k}^{i,j}|,|b_k^{i}|,|a_i|\leq \mathcal R,\nonumber\\
      &&
      1\leq i\leq d_{k},1\leq j\leq d_{k-1},
        1\leq k\leq
      L\},
\end{eqnarray}
where
\begin{equation}\label{Def.R}
   \mathcal R:= C_3\max\left\{\frac{2N^d(N^*)^{1+pr}}{s}, (N^*)^{2\gamma d+\gamma r}\left(\frac{N^d}{s}\right)^{\gamma/p}\right\}.
\end{equation}
Then, it is easy to verify that
\begin{equation}\label{subn3}
  N_{3,N^*,\tau}\in  H_{n,L,\mathcal R}
\end{equation}
with $\tau=\frac{s}{2N^d(N^*)^{1+pr}}$.

We  consider the
  generalization error estimates for implementing ERM   on $\mathcal H_{n,L,\mathcal R}$ as follows:
\begin{equation}\label{ERM}
      f_{D,n,L}:={\arg\min}_{f\in \mathcal
      H_{n,L,\mathcal R}}\frac1m\sum_{i=1}^m[f(x_i)-y_i]^2.
\end{equation}
Since $|y_i|\leq M$, it is natural to project the final output
$f_{D,n,L}$ to the interval $[-M, M]$ by the truncation operator
$\pi_M
f_{D,n,L}(x):=sign(f_{D,n,L}(x))\min\{|f_{D,n,L}(x)|,M\}.$

 Let $p\geq 2$ and $J_p$ be the identity mapping
$$
       L^p(\mathcal X)       ~~ {\stackrel{J_p}{\longrightarrow}}~~  L^2_{\rho_X}.
$$
and $D_{\rho _{X},p}=\|J_p\|.$  Then $D_{\rho _{X},p}$ is called the
distortion of $\rho _{X}$ (with respect to the Lebesgue measure)
\cite{Zhou2006}, which measures how much $\rho _{X}$ distorts the
Lebesgue measure. In our analysis, we assume $D_{\rho _{X},p}<\infty$, which holds for
 the uniform
distribution for all $p\geq2$ obviously. According
to the definition, for each $f\in L_{\rho_X}^2\cap L^p(\mathbb I^d)$, we have
\begin{equation}\label{distrotion}
    \|f\|_\rho \leq   D_{\rho _{X},p}\|f\|_{L^p(\mathcal
    X)}.
\end{equation}
The following theorem  with  proof to be given in Section \ref{Sec.Lower bound}  shows that the simple ERM strategy (\ref{ERM}) based on deep ReLU nets has the capability of realizing the optimal learning rates established in  Theorem \ref{Theorem:sampling-theorem}.

\begin{theorem}\label{Theorem: ERM}
Let $f_{D,n,L}$ be defined by (\ref{ERM}) with $L,n$ satisfying (\ref{hypoth1})  and $\mathcal R$ satisfying (\ref{Def.R}).
Suppose that
\begin{equation}\label{def.N*}
       (N^*)^{2r+d}\sim
        m\left(\frac{s}{N^d}\right)^\frac{2}p/\log m,
\end{equation}
 and
 \begin{equation}\label{restions-on-m-1}
 \frac{m}{\log m}\geq C^*\frac{N^\frac{2d+2rp+dp}{(2r+d)p}}{s^\frac{2}{2rp+dp}}.
 \end{equation}
Then
\begin{eqnarray}\label{almost optimal learning rate}
          &&C_1 m^{-\frac{2r}{2r+dd}}\left(\frac{s}{N^d}\right)^{\frac{d}{2r+d}} \nonumber\\
            &\leq&
         \sup_{f_\rho\in Lip^{(N,s,r,c_0)}}\mathbf E\left\{\mathcal E(\pi_Mf_{D,n,L^*})-\mathcal
         E(f_\rho)\right\}  \nonumber\\
           &\leq&
         C_6  \left(\frac{m}{\log m}\right)^\frac{-2r}{2r+d}  \left(\frac{s}{N^d}\right)^{\frac2p-\frac{2r}{2r+d}},
\end{eqnarray}
 where  $C_1$, $C_6 $ are constants independent of $m$, $s$, or $N$ and $a\sim b$ with $a,b\geq0$ denotes that there exist positive absolute constants $\hat{C}_1,\hat{C}_2$ such that $\hat{C}_1 a\leq b\leq \hat{C}_2 a$.
\end{theorem}

If $\rho_X$ is the uniform distribution, then (\ref{distrotion}) holds with $p=2$ and $D_{\rho_X,p}=1$. Hence, if $f_\rho\in Lip^{(N,s,r,c_0)}$, we have
\begin{eqnarray}\label{almost optimal learning ratewith uniform}
         &&\mathbf E\left\{\mathcal E(\pi_Mf_{D,n,L^*})-\mathcal
         E(f_\rho)\right\} \nonumber\\
            &\leq&
         C_6  m^{-\frac{2r}{2r+d}} (\log m)^{2r/(r+d)} \left(\frac{s}{N^d}\right)^{\frac{d}{2r+d}},
\end{eqnarray}
which coincides with the optimal learning rates in Theorem \ref{Theorem:sampling-theorem} up to a logarithmic factor. Theorem \ref{Theorem: ERM} thus presents a theoretical verification on the success of deep learning in spatial sparseness related applications for  massive data. In particular, it presents an intuitive explanation on why deep learning performs so well in   handwritten digit recognition \cite{Chherawala2016}. As shown in Figure \ref{Figure:sparse-and-partition}, high-resolution of a figure implies large size of data, which admits an extremely large partitions for the input space with small sparsity $s$. Then, the additional term $\left(\frac{s}{N^d}\right)^{\frac{d}{2r+d}}$ in   Theorem \ref{Theorem: ERM} yields a small generalization error.

Learning spatially sparse and smooth functions was first studied in \cite{Lin2018} and similar learning rate as that in Theorem \ref{Theorem: ERM} has been derived. In comparison with \cite{Lin2018}, there are three novelties of our work. The first is that we deduce   lower bounds for learning these functions and show the optimality for the derived learning rates, while \cite{Lin2018} only focused on upper bounds. The second is that the range of $r$ in our study is $r>0$, while that in \cite{Lin2018} is $0<r\leq 1$. In view of the discussion in  \cite{Yarotsky2017}, the depth is necessary for extending the range from $0<r\leq 1$ to $r>0$. Thus, more layers  are required in our analysis to show the advantage of deep nets. We would like to point out that the activation function in the present paper is the widely used ReLU function, while the activation functions in \cite{Lin2018} are hybrid functions including the Heaviside function in the first layer and continuous sigmoid-type function in other layers.

\section{Upper Bound Estimates}\label{Sec.Upperbound}

This section is devoted to the proof of Proposition \ref{Proposition:Local approximation}, Proposition \ref{Proposition:Approximation error for large r}, and the upper bounds in (\ref{almost optimal learning rate}) and (\ref{samplingtheorem}). It should be noted that the upper bound in (\ref{samplingtheorem}) is a direct corollary of the upper of (\ref{almost optimal learning rate}), with $p=2$.

\subsection{Proofs of Proposition \ref{Proposition:Local approximation}}
\begin{IEEEproof}[Proof of Proposition \ref{Proposition:Local approximation}] For $x\in\mathbb I^d$, it follows from
(\ref{Deteailed trapezoid}) that $0\leq T_{\tau,a,b}(x^{(j)})\leq 1$
for any $j\in\{1,\dots,d\}$. This implies that
$\sum_{j=1}^dT_{\tau,a,b,}(x^{(j)})\leq d$ and consequently $0\leq
N_{a,b,\tau}(x)\leq 1.$ If $x\notin[a-\tau,b+\tau]^d$, there is at least one
$j_0\in\{1,\dots,d\}$ such that $x^{(j_0)}\notin[a-\tau,b+\tau]$.
This together with (\ref{Deteailed trapezoid}) shows that
$T_{\tau,a,b}(x^{(j_0)})=0$. Therefore
$\sum_{j=1}^dT_{\tau,a,b}(x^{(j)})\leq d-1$, which implies
$N_{a,b,\tau}(x)=0$. If $x\in[a,b]^d$, then $x^{(j)}\in[a,b]$ for
every $j\in\{1,\dots,d\}$. Hence, it follows from (\ref{Deteailed trapezoid}) that
$T_{\tau,a,b}(x^{(j)})=1$ for every $j \in\{1,\dots,d\}$, which
implies that $\sum_{j=1}^dT_{\tau,a,b}(x^{(j)})=d$ and
$N_{a,b,\tau}(x)=1$. This completes the proof of Proposition
\ref{Proposition:Local approximation}.
\end{IEEEproof}

\subsection{Proof of Proposition \ref{Proposition:Approximation error for large r}}

Before presenting the proof of Proposition \ref{Proposition:Approximation error for large r}, we need several lemmas.
The first one can be found in \cite[Lemma 1]{Kohler2014}.
\begin{lemma}\label{Lemma:Taylor polynomials}
Let $r=u+v$ with $u\in\mathbb N_0$ and $0<v\leq 1$. If $f\in
Lip^{(r,c_0)}$,  $x_0\in\mathbb R^d$  and $p_{u,x_0,f}$ is the
Taylor polynomial of $f$ with degree $u$ at    $x_0$,
i.e.,
\begin{eqnarray}\label{Taylor polynomial}
    p_{u,x_0,f}(x)&=&\sum_{k_1+\cdots+k_d\leq u}\frac{1}{k_1!\dots
    k_d!}\frac{\partial^{k_1+\dots+
    k_d}f(x_0)}{\partial^{k_1}x^{(1)}\dots\partial^{k_d}x^{(d)}}\nonumber\\
    &&(x^{(1)}-x_0^{(1)})^{k_1}
    \cdots(x^{(d)}-x_0^{(d)})^{k_d},
\end{eqnarray}
then
\begin{equation}\label{Taylor approximation}
   |f(x)-p_{u,x_0,f}(x)|\leq c_1\|x-x_0\|^r,\qquad\forall\ x\in\mathbb
   I^d,
\end{equation}
where $c_1$ is a constant depending only on $r$, $c_0$ and $d$.
\end{lemma}

For $\tau>0$,
define the localized Taylor polynomials by
\begin{equation}\label{Def.N3}
     N_{2,N^*,\tau}(x):=\sum_{k=1}^{(N^*)^d}p_{u,\xi_k,f}(x)N_{1,N^*,
     \xi_{k},\tau}(x),
\end{equation}
where $N_{1,N^*,
     \xi_{k},\tau}$ is given in (\ref{NN-for-localization}).
In the following lemma, we
present an upper bound estimate for approximating functions in
$Lip^{(N,s,r,c_0)}$ by $N_{2,N^*,\tau}$.

\begin{lemma}\label{Lemma:Approximation error for product}
Let $1\leq p<\infty$ and $N^*\geq 4N$. If $f\in Lip^{(N,s,r,c_0)}$
with $N,s\in\mathbb N$, $r>0$ and $c_0>0$, then for any
 $0<\tau \leq
 \frac{s}{2N^d(N^*)^{1+pr}}$, it follows that
\begin{equation}\label{app theorem for probduct1}
   \|f-N_{2,N^*,\tau}\|_{L^p(\mathbb I^d)}
   \leq c_2(N^*)^{-r}\left(\frac{s}{N^d}\right)^{1/p}
\end{equation}
and
\begin{equation}\label{bound1-for-product}
    \|N_{2,N^*,\tau}\|_{L^\infty(\mathbb I^d)}\leq c_3,
\end{equation}
where $c_2,c_3$ are constants  dependent only on $d$, $r$, $c_0$ and $\|f\|_{L^\infty(\mathbb I^d)}$.
\end{lemma}

\begin{IEEEproof} Observe that  $\mathbb I^d=\bigcup_{k=1}^{(N^*)^d}B_k$. Then for
each $x\in\mathbb I^d$, let $k_x$ be  the smallest $k$ such that $x \in B_{k_x}$ Note that $k_x$ is unique (the last restriction is for those points $x$ on boundaries of cubes $B_k$).  It follows from
(\ref{Def.N3}) that
\begin{eqnarray*}
      &&f(x)-N_{2,N^*,\tau}(x)\\
       &=&
      f(x)-\sum_{k=1}^{(N^*)^d}p_{u,\xi_k,f}(x)N_{1,N^*,
     \xi_{k},\tau}(x)   \\
     &=&
     f(x)-p_{u,\xi_{k_x},f}(x)-\sum_{k\neq k_x}p_{u,\xi_k,f}(x)N_{1,N^*,
     \xi_{k},\tau}(x)\\
     &+&
     p_{u,\xi_{k_x},f}(x)[1-N_{1,N^*,
     \xi_{k_x},\tau}(x)].
\end{eqnarray*}
But (\ref{property localization}) implies that
$1-N_{1,N^*,
     \xi_{k_x},\tau}(x)=0$. Thus,
\begin{eqnarray}\label{app decomp111}
   &&\|f-N_{2,N^*,\tau}\|_{L^p(\mathbb I^d)}\leq \left\|f-p_{u,\xi_{k_x},f}\right\|_{L^p(\mathbb I^d)} \nonumber \\
   &+&\left\|\sum_{k\neq k_x}p_{u,\xi_{k},f}(x)N_{1,N^*,
     \xi_{k},\tau}(x)\right\|_{L^p(\mathbb I^d)}.
\end{eqnarray}
We first estimate the first term on the right-hand side of (\ref{app decomp111}).
For  $j\in\Lambda_s$, define
\begin{equation}\label{Lambda tilde}
   \tilde{\Lambda}_{j}:=\{k\in\{1,\dots,(N^*)^d\}: B_k\cap A_j\neq \varnothing\}.
\end{equation}
Since $\{A_j\}_{j=1}^{N^d}$ and $\{B_k\}_{k=1}^{(N^*)^d}$ are cubic
partitions of $\mathbb I^d$ and $N^*\geq 4N$, we have
\begin{equation}\label{Card lambda tilde}
     |\tilde{\Lambda}_j|\leq \left(\frac{N^*}{N}+2\right)^d\leq
     \left(\frac{2N^*}{N}\right)^d,\qquad\ \forall j\in\Lambda_s.
\end{equation}
In view of  (\ref{Lambda tilde}), we obtain
\begin{equation}\label{setcap}
        \mathbb I^d\subseteq
        \left[\bigcup_{j\in\Lambda_s}\left(\bigcup_{k\in\tilde{\Lambda}_j}B_k\right)\right]
         \bigcup
         \left[\left(\bigcup_{k\in\{1,\dots,(N^*)^d\}\backslash(\cup_{j\in\Lambda_s}\tilde{\Lambda}_j)}B_k\right)
         \right].
\end{equation}
Then,
\begin{eqnarray}\label{decomposition 2 part111}
   &&\left\|f-p_{u,\xi_{k_x},f}\right\|_{L^p(\mathbb I^d)}
   =
   \int_{\mathbb I^d}\left|f(x)-p_{u,\xi_{k_x},f}(x)\right|^pdx\nonumber\\
   &\leq&
   \left[\sum_{j\in\Lambda_s}\sum_{k\in\tilde{\Lambda}_j}
   +
   \sum_{k\in\{1,\dots,(N^*)^d\}\backslash(\cup_{j\in\Lambda_s}\tilde{\Lambda}_j)}\right]\nonumber\\
   &&\int_{B_k}
   \left|f(x)-p_{u,\xi_{k_x},f}(x)\right|^pdx.
\end{eqnarray}
From (\ref{Lambda tilde}) again, for any $k\in
\{1,\dots,(N^*)^d\}\backslash(\cup_{j\in\Lambda_s}\tilde{\Lambda}_j)$,
we have  $B_k\cap S=\varnothing$, which together with (\ref{Taylor polynomial}) and  $f\in
Lip^{(N,s,r,c_0)}$   yields  $f(x)=p_{u,\xi_{k_x},f}(x)=0$ for
$x\in
  B_k$. Hence,
\begin{equation}\label{est a2111}
        \sum_{k\in\{1,\dots,(N^*)^d\}\backslash(\cup_{j\in\Lambda_s}
        \tilde{\Lambda}_j)}\int_{B_k}\left|f(x)-p_{u,\xi_{k_x},f}(x)\right|^pdx=0.
\end{equation}
Since   $f\in Lip^{(N,s,r,c_0)}$, it follows from Lemma \ref{Lemma:Taylor polynomials}   and (\ref{Card lambda tilde})     that
\begin{eqnarray}\label{est a1111}
   &&
   \sum_{j\in\Lambda_s}\sum_{k\in\tilde{\Lambda}_j}\int_{B_k}\left|f(x)-p_{u,\xi_{k_x},f}(x)\right|^pdx\nonumber\\
   &\leq&
   c_1^p\sum_{j\in\Lambda_s}\sum_{k\in\tilde{\Lambda}_j}\int_{B_k} \|x-\xi_{k_x}\|^{pr}dx\nonumber\\
   &\leq&
   c^p_12^dd^{pr/2}(N^*)^{-pr}\frac{s}{N^d}.
\end{eqnarray}
Inserting (\ref{est a1111}) and (\ref{est a2111}) into (\ref{decomposition
2 part111}), we obtain
\begin{equation}\label{app first term111}
         \left\|f-p_{u,\xi_{k_x},f}\right\|_{L^p(\mathbb I^d)}
         \leq
         c_12^{d/p}d^{r/2}(N^*)^{-r}\left(\frac{s}{N^d}\right)^{1/p}.
\end{equation}
Now we estimate the second term of the right-hand side of (\ref{app
decomp111}). For each $k'\in\{1,\dots,(N^*)^d\}$, define
\begin{equation}\label{Def: xi}
      \Xi_{k'}:=\{k\in\{1,\dots,(N^*)^d\}:\tilde{B}_{k,\tau}\cap B_{k'}\neq\varnothing,k\neq k'\}.
\end{equation}
Since $0<\tau  \leq \frac1{2N^*}$, it is easy to verify that
\begin{equation}\label{card xi}
        |\Xi_{k'}|\leq 3^d-1,\qquad \forall\ k'\in \{1,\dots,(N^*)^d\}.
\end{equation}
Noting further that
\begin{eqnarray}\label{a 1111}
    &&\left\|\sum_{k\neq k_x}p_{u,\xi_{k},f}(x)N_{1,N^*,
     \xi_{k},\tau}(x)\right\|^p_{L^p(\mathbb I^d)}\nonumber\\
  &\leq&\sum_{k'=1}^{(N^*)^d}
   \int_{B_{k'}}\left|\sum_{k\neq k_x}p_{u,\xi_{k},f}(x)N_{1,N^*,
     \xi_{k},\tau}(x)\right|^pdx,
\end{eqnarray}
we obtain, from (\ref{Def: xi}), (\ref{card xi}),
  (\ref{property
localization}) and $|N_{1,N^*, \xi_{k},\tau}(x)|\leq 1$, that
\begin{eqnarray*}
    &&\int_{B_{k'}}\left|\sum_{k\neq k_x}p_{u,\xi_{k},f}(x)N_{1,N^*,
     \xi_{k},\tau}(x)\right|^pdx\\
     &=&
     \int_{B_{k'}}\left|\sum_{k\in\Xi_{k'}}p_{u,\xi_{k},f}(x)N_{1,N^*,
     \xi_{k},\tau}(x)\right|^pdx\\
     &\leq&
   \max_{1\leq k\leq(N^*)^d}\|p_{u,\xi_{k},f}\|^p_{L^\infty(\mathbb I^d)} \\
   &\times& \sum_{\ell\in\Xi_{k'}}\int_{\tilde{B}_{\ell,\tau}\cap B_{k'}}\left|\sum_{k\in\Xi_{k'}}N_{1,N^*,
     \xi_{k},\tau}(x)\right|^pdx\\
     &\leq&
     3^{dp}\max_{1\leq k\leq(N^*)^d}\|p_{u,\xi_{k},f}\|^p_{L^\infty(\mathbb
     I^d)}\sum_{\ell\in\Xi_{k'}}\int_{\tilde{B}_{\ell,\tau}\cap B_{k'}}dx.
\end{eqnarray*}
But
 $k'\notin \Xi_{k'}$  implies that for any $
       \ell\in\Xi_{k'}$,
\begin{equation}\label{Volume1}
       \int_{\tilde{B}_{\ell,\tau}\cap B_{k'}}dx\leq(1/N^*+2\tau)^d-(1/N^*)^d\leq
       2d\tau(N^*)^{1-d},
\end{equation}
where  the mean value theorem is applied to yield  the last inequality.
Thus,
\begin{eqnarray*}
       &&\int_{B_{k'}}\left|\sum_{k\neq k_x}p_{u,\xi_{k},f}(x)N_{1,N^*,
     \xi_{k},\tau}(x)\right|^pdx\\
     &\leq&
     2d3^{d(p+1)}\max_{1\leq k\leq(N^*)^d}\|p_{u,\xi_{k},f}\|^p_{L^\infty(\mathbb I^d)}\tau(N^*)^{1-d}.
\end{eqnarray*}
Plugging the above estimate into (\ref{a 1111}), we may conclude from
$0<\tau\leq (N^*)^{-1-pr}\left(\frac{s}{2N^d}\right)$ that
\begin{eqnarray}\label{app second term111}
       &&\left\|\sum_{k\neq k_x} p_{u,\xi_{k},f}(x)  N_{1,N^*,
     \xi_{k},\tau}(x)\right\|_{L^p(\mathbb I^d)}\\
   & \leq& d^{1/p}3^{2d}\max_{1\leq k\leq(N^*)^d}\|p_{u,\xi_{k},f}\|_{L^\infty(\mathbb I^d)}(N^*)^{-r}\left(\frac{s}{N^d}\right)^{\frac1p}.\nonumber
\end{eqnarray}
Inserting (\ref{app first term111}) and (\ref{app second term111}) into
(\ref{app decomp111}) and noting that
$$
   \max_{1\leq k\leq(N^*)^d}\|p_{u,\xi_{k},f}\|_{L^\infty(\mathbb I^d)}\leq
  \|f\|_{L^\infty(\mathbb I^d)}+c_1d^{r/2},
$$
from (\ref{Taylor approximation}), we deduce that
$$
   \|f-N_{2,N^*,\tau}\|_{L^p(\mathbb I^d)}
   \leq c_2(N^*)^{-r}\left(\frac{s}{N^d}\right)^{1/p},
$$
with $
       c_2:= c_12^{d/p}d^{r/2}+d^{1/p}3^{2d}(\|f\|_{L^\infty(\mathbb I^d)}+c_1d^{r/2}).$
This proves (\ref{app theorem for probduct1}).

We now turn to prove (\ref{bound1-for-product}). First, (\ref{Def.N3}) and (\ref{property localization}) imply that for   $x\in\mathbb I^d$,
\begin{eqnarray*}
     &&N_{2,N^*,\tau}(x)=\sum_{k=1}^{(N^*)^d}p_{u,\xi_k,f}(x)N_{1,N^*,
     \xi_{k},\tau}(x)\\
     &=&
     \sum_{k:\tilde{B}_{k,\tau}\cap B_{k_x}\neq\varnothing}
     p_{u,\xi_k,f}(x)N_{1,N^*,
     \xi_{k},\tau}(x).
\end{eqnarray*}
Since $0<\tau\leq 1/(2N^*)$, it follows from (\ref{card xi}) and $0\leq N_{1,N^*,
     \xi_{k},\tau}(x)\leq 1$ that
$$
       |N_{2,N^*,\tau}(x)|\leq 3^d(\|f\|_{L^\infty(\mathbb I^d)}+c_1d^{r/2})=:c_3,\qquad\forall x\in\mathbb I^d.
$$
  This completes the proof of Lemma \ref{Lemma:Approximation error for product}.
\end{IEEEproof}

The following ``product-gate'' property for deep ReLU nets
can be found in \cite{Han2019}.

\begin{lemma}\label{lemma:product gate 2}
Let $\theta>0$  and $\tilde{L}\in\mathbb N$ with $\tilde{L}>(2\theta)^{-1}$. For any $\ell\in\{2,3,\dots\}$ and $\varepsilon\in
(0,1)$, there exists a deep ReLU net $\tilde{\times}_\ell$
with $2\ell \tilde{L}+8\ell$
layers and at most $c_4\ell^{\theta} \varepsilon^{-\theta}$ free
  parameters bounded by $\ell^\gamma\varepsilon^{-\gamma}$,
 such that
$$
       |t_1t_2\cdots t_\ell-\tilde{\times}_\ell(t_1,\dots,t_\ell)|\leq
       \varepsilon,\quad \forall t_1,\dots,t_\ell\in[-1,1],
$$
where $c_4$ and $\gamma$ are   constants  depending only on $\theta$ and $\tilde{L}$.
\end{lemma}

 For $\beta\in\mathbb N_0$ and $B>0$, define
$$
     \mathcal P_{\beta,B}^d:=\left\{\sum_{|\alpha|\leq \beta}c_\alpha
     x^\alpha:|c_\alpha|\leq B\right\},
$$
where
$\alpha=(\alpha_1,\dots,\alpha_d)\in\mathbb N_0^d$,
$|\alpha|=\alpha_1+\dots+\alpha_d$ and
$x^\alpha=(x^{(1)})^{\alpha_1}\cdots(x^{(d)})^{\alpha_d}$. The following lemma was proved in \cite{Han2019}.

\begin{lemma}\label{lemma:polynomial}
Let $\beta\in\mathbb N_0$, $B>0$, $\theta>0$  and $\tilde{L}\in\mathbb N$ with $\tilde{L}>(2\theta)^{-1}$.   For every $P\in\mathcal P_{\beta,B}^d$ and
$0<\varepsilon<1$, there is   a deep ReLU net structure
 with     $2\beta \tilde{L}+8\beta+1$ layers and at most $\beta^d+c_4(\beta^{d+1} B)^{\theta}
\varepsilon^{-\theta}$ free parameters   bounded by
$\max\{B,(\beta^{d+1} B)^\gamma\varepsilon^{-\gamma}\}$ such that for any $P\in\mathcal P^d_{\beta,B}$ there exists a deep ReLU net $h_P$ with the aforementioned structure that satisfies
$$
       |P(x)-h_P(x)|\leq \varepsilon,\qquad\forall x\in \mathbb I^d.
$$
\end{lemma}

Let $c_5$ be a constant that satisfies
\begin{eqnarray*}
    &&c_42^{\frac{d}{2r+d}}c_5^{-\frac{d}{2r+d}}+ 4d+1+r^d \\
    &+&c_4(u^{d+1}(\|f\|_{L^\infty(\mathbb I^d)}+c_1d^{r/2}))^{\frac{d}{r+d}}c_5^{-\frac{d}{r+d}}\\
     &\leq&  2(4d+1+r^d).
\end{eqnarray*}
For $N^*>\max\left\{c_5^{1/(d+r)},1\right\}$,
let $\tilde{\times}_2$ be the deep net as introduced in Lemma \ref{lemma:product gate 2} with $\ell=2$,  $\theta=\frac{d}{2d+r}$, $\varepsilon=c_5(N^*)^{-2d-r}$ and let  $\tilde{L}=\lceil 2+r/d\rceil$. Denote by  $h_{p_{u,\xi_{k},f}}$ the deep ReLU net in Lemma \ref{lemma:polynomial} with $P=p_{u,\xi_{k},f}$, $\beta=u$, $B=\|f\|_{L^\infty(\mathbb I^d)}+c_1d^{r/2}$,
$\varepsilon=c_5(N^*)^{-r-d}$, $\theta=d/(r+d)$, and $\tilde{L}=\lceil 1+r/d\rceil$.
From Lemma \ref{lemma:polynomial}, we have, for any  $x\in\mathbb I^d, k=1,\dots,(N^*)^d$, that
\begin{equation}\label{boundforhp}
   |h_{p_{u,\xi_{k},f}}(x)|\leq |p_{u,\xi_{k},f}(x)|+1\leq  \|f\|_{L^\infty(\mathbb I^d)}+c_1d^{r/2} +1.
\end{equation}
 Next consider
\begin{eqnarray}\label{Def.N31}
     &&N_{3,N^*,\tau}(x):=(\|f\|_{L^\infty(\mathbb I^d)}+c_1d^{r/2}+1) \\
     &\times&\sum_{k=1}^{(N^*)^d}\tilde{\times}_2\left(\frac{h_{p_{u,\xi_{k},f}}(x)}{\|f\|_{L^\infty(\mathbb I^d)}+c_1d^{r/2}+1},N_{1,N^*,
     \xi_{k},\tau}(x)\right). \nonumber
\end{eqnarray}
Noting that the parameters of the deep nets $\tilde{\times}_2(t_1,t_2)$ are independent of  $t_1,t_2\in[-1,1]$, we may conclude that
$N_{3,N^*,\tau}$ is  a deep net with    $\lceil 25+4r/d+2r^2/d+10r\rceil$ layers and at most
 $C_1^*(N^*)^d$ free  parameters  with $C_1^*:=2(4d+1+r^d)$
 that are bounded by  $\tilde{B}^*$ defined by (\ref{boundofweights}) with $C_3:=2r^{d+1} (\|f\|_{L^\infty(\mathbb I^d)}+c_1d^{r/2})$.
With these preparations, we can now prove Proposition \ref{Proposition:Approximation error for large r} as follows.

\begin{IEEEproof}[Proof of Proposition \ref{Proposition:Approximation error for large r}]
 By applying the triangle inequality, we have
\begin{eqnarray}\label{triangle111}
    &&\|f-N_{3,N^*,\tau}\|_{L^p(\mathbb I^d)}\nonumber\\
    &\leq&
    \left\|N_{2,N^*,\tau}(x)-\sum_{k=1}^{(N^*)^d}h_{p_{u,\xi_{k},f}}(x)N_{1,N^*,
     \xi_{k},\tau}(x)\right\|_{L^p(\mathbb I^d)}\nonumber\\
    & +&
     \left\|\sum_{k=1}^{(N^*)^d}h_{p_{u,\xi_{k},f}}(x)N_{1,N^*,
     \xi_{k},\tau}(x)-N_{3,N^*,\tau}(x)\right\|_{L^p(\mathbb I^d)} \nonumber\\
     &+&
     \|f-N_{2,N^*,\tau}(x)\|_{L^p(\mathbb I^d)}\nonumber\\
     &=:&
     I_1+I_2+I_3.
\end{eqnarray}
It follows from $p\geq 1$, $N^*\geq 4N$ and  Lemma \ref{lemma:product gate 2}  with $\ell=2$, $\theta=\frac{d}{2d+r}$, $\varepsilon=c_5(N^*)^{-2d-r}$ and $\tilde{L}=\lceil 2+r/d\rceil$ that
\begin{eqnarray*}
       I_2
       &\leq& c_5(\|f\|_{L^\infty(\mathbb I^d)}+c_1d^{r/2}+1)(N^*)^{-r-d}\\
       &\leq& c_5(\|f\|_{L^\infty(\mathbb I^d)}+c_1d^{r/2}+1)(N^*)^{-r}\left(\frac{s}{N^d}\right)^{1/p}.
\end{eqnarray*}
Similarly,  we also note $0\leq N_{1,N^*, \xi_{k},\tau}(x)\leq 1$ and Lemma \ref{lemma:polynomial}  with  $\beta=u$, $B= \|f\|_{L^\infty(\mathbb I^d)}+c_1d^{r/2}$,
$\varepsilon=c_5(N^*)^{-r-d}$, $\theta=d/(r+d)$, and $\tilde{L}=\lceil 1+r/d\rceil$ imply
\begin{eqnarray*}
       I_1
       &\leq&
        c_5(N^*)^{-r-d}\left(\int_{\mathbb I^d}\left|\sum_{k=1}^{(N^*)^d}N_{1,N^*,
     \xi_{k},\tau}(x)\right|^pdx\right)^{1/p}\\
     &=& c_5(N^*)^{-r-d}\left(\int_{\mathbb I^d}\left|\sum_{k\neq k_x}N_{1,N^*,
     \xi_{k},\tau}(x)\right|^pdx\right)^{1/p}\\
     & +&
     c_5(N^*)^{-r-d}.
\end{eqnarray*}
The same approach as that in the proof of (\ref{app second term111}) yields that, for $0<\tau\leq (N^*)^{-1-pr}\left(\frac{s}{2N^d}\right)$,
$$
    \left(\int_{\mathbb I^d}\left|\sum_{k\neq k_x}N_{1,N^*,
     \xi_{k},\tau}(x)\right|^pdx\right)^{1/p}
     \leq d^{1/p}3^{2d}(N^*)^{-r}\left(\frac{s}{N^d}\right)^{\frac1p}.
$$
Therefore,
$$
    I_1\leq  c_6(N^*)^{-r-d} \leq c_6(N^*)^{-r}\left(\frac{s}{N^d}\right)^{\frac1p},
$$
where $c_6:=c_5(1+d^{1/p}3^{2d})$.
Furthermore, by Lemma \ref{Lemma:Approximation error for product} that under
 $0<\tau \leq
 \frac{s}{2N^d(N^*)^{1+pr}}$, we obtain
$$
   I_3
   \leq c_2(N^*)^{-r}\left(\frac{s}{N^d}\right)^{1/p}.
$$
Plugging the estimates of $I_1,I_2,I_3$ into (\ref{triangle111}),
we then have
$$
     \|f-N_{3,N^*,\tau}\|_{L^p(\mathbb I^d)}
     \leq C_4(N^*)^{-r}\left(\frac{s}{N^d}\right)^{1/p}
$$
with
$C_4:=c_2+c_6+c_5$. Thus, (\ref{app theorem for large r}) holds.

Now we turn to the proof of (\ref{bound-N3-NN}). According to (\ref{Def.N31}) and  Lemma \ref{lemma:product gate 2}  with $\ell=2$, $\theta=\frac{d}{2d+r}$, $\varepsilon=(N^*)^{-2d-r}$ and $\tilde{L}=\lceil 2+r/d\rceil$, we have
\begin{eqnarray*}
     &&|N_{3,N^*,\tau}(x)|\leq \left|\sum_{k=1}^{(N^*)^d}h_{p_{u,\xi_{k},f}}(x)N_{1,N^*,
     \xi_{k},\tau}(x)\right|\\
     &+& c_5(N^*)^{-d-r}(\|f\|_{L^\infty(\mathbb I^d)}+c_1d^{r/2}+1).
\end{eqnarray*}
But (\ref{property localization}), together with $0<\tau\leq 1/(2N^*)$, (\ref{card xi}), $0\leq N_{1,N^*,
     \xi_{k},\tau}(x)\leq 1$ and (\ref{boundforhp}) yields
$$
    \left|\sum_{k=1}^{(N^*)^d}h_{p_{u,\xi_{k},f}}(x)N_{1,N^*,
     \xi_{k},\tau}(x)\right|\leq 3^d(\|f\|_{L^\infty(\mathbb I^d)}+c_1d^{r/2}+1),
$$
which implies (\ref{bound-N3-NN}) with $C_5:=(c_5+3^d)(\|f\|_{L^\infty(\mathbb I^d)}+c_1d^{r/2}+1) $.
 This completes the proof of Proposition \ref{Proposition:Approximation error for large r}.
 \end{IEEEproof}

\subsection{Proof of Theorem \ref{Theorem: ERM}}

Let $\mathbb B$ be a Banach space and $V$ be  a subset of $\mathbb B$. Denote by $\mathcal N(\varepsilon,V,\mathbb B)$ the $\varepsilon$-covering number
\cite[Chap. 9]{Gyorfi2002}
of $V$ under the metric of $\mathbb B$, which is
the minimal number of elements in an $\varepsilon$-net of $V$.
The following lemma    proved in \cite[Theorem 1]{Guo2019} gives rise to  a  tight estimate for the covering number of deep ReLU nets.

\begin{lemma}\label{Lemma:covering number}
Let $\mathcal H_{n,L, \mathcal R}$ be defined by
(\ref{Hypothesis space}). Then
\begin{eqnarray}\label{covering1}
  \mathcal N\left( \varepsilon,\mathcal H_{n,L,\mathcal
       R},L^\infty(\mathbb I^d)\right)
     \leq
   \left(c_7\mathcal RD_{\max}\right)^{3(L+1)^2n}\varepsilon^{-
   n},
\end{eqnarray}
where  $c_7$ is a
constant depending only on $d$ and $D_{\max}=\max_{0\leq \ell\leq L}d_\ell$.
\end{lemma}

To prove Theorem \ref{Theorem: ERM}, we also  need the following
lemma, the proof of which
can be found in \cite{Chui2018a}.

\begin{lemma}\label{Lemma:oracle}
 Let $\mathcal H$ be a
collection of functions defined on $\mathbb I^d$ and define
\begin{equation}\label{ERM!!!!!!}
        f_{D,\mathcal H}=\arg\min_{f\in\mathcal H}
        \frac1m\sum_{i=1}^m(f(x_i)-y_i)^2.
\end{equation}
Suppose there exist  $n', \mathcal U>0$, such that
\begin{equation}\label{oracle condition}
    \log \mathcal N(\varepsilon,\mathcal H,L^\infty(\mathbb I^d))\leq
     n'\log \frac{\mathcal U}{\varepsilon},\qquad\forall  \varepsilon>0.
\end{equation}
Then for any $h\in\mathcal H$ and $\varepsilon>0$,
\begin{eqnarray*}
      &&  Pr\{\|\pi_Mf_{ D,\mathcal
      H}-f_\rho\|_\rho^2>\varepsilon+2\|h-f_\rho\|_\rho^2\}\\
      &\leq&
      \exp\left\{n'\log\frac{16\mathcal UM}{\varepsilon}-\frac{3m\varepsilon}{512M^2}\right\}\\
      &+&
      \exp\left\{\frac{-3m\varepsilon^2}{16(3M+\|h\|_{L_\infty(\mathcal X)})^2\left(
    6\|h-f_\rho\|_\rho^2+
    \varepsilon\right)}\right\}.
\end{eqnarray*}
\end{lemma}

Now we are in a position to prove the upper bound of (\ref{almost optimal learning rate}).

\begin{IEEEproof}[Proof of the upper bound of (\ref{almost optimal learning rate})] For $N^*\geq \max\{4N,\tilde{C}\}$,
Proposition \ref{Proposition:Approximation error for large r} implies that
there exists an
 $h_\rho\in \mathcal H_{L,n,\mathcal R}$  with $L,n$ satisfying (\ref{hypoth1}) and
$\mathcal R$ satisfying  (\ref{Def.R})
such that
\begin{eqnarray*}
      &&\|f_\rho-h_{\rho}\|^2_\rho\leq D_{\rho_X,p}^2\|f_\rho-h_{\rho}\|^2_{L^p(\mathbb I^d)}\\
       &\leq& C_4^2D_{\rho_X,p}^2(N^*)^{-2r}\left(\frac{s}{N^d}\right)^{2/p}=:\mathcal A_{p}.
\end{eqnarray*}
Recalling (\ref{bound-N3-NN}), we have
$$
       \|h_{\rho}\|_{L^\infty(\mathbb I^d)} \leq C_5.
$$
But Lemma \ref{Lemma:covering number}, together  with the structure of deep nets in Proposition \ref{Proposition:Approximation error for large r}, (\ref{hypoth1}) and (\ref{Def.R}), implies $D_{\max}\leq c_8 n$ with $c_8$ depending only on $r$ and $d$ and
\begin{eqnarray*}
   &&\log\mathcal N\left( \varepsilon,\mathcal H_{n,L,\mathcal
       R},L^\infty(\mathbb I^d)\right)
     \leq
       c_9L^2n\log\frac{\mathcal Rn}{\varepsilon}\\
       &\leq&
       c_{10}(N^*)^d\log\frac{N^*N^d}{s\varepsilon}
\end{eqnarray*}
for some positive constants $c_9,c_{10}$ depending only on $d,r,C_1^*,c_7,c_8,\gamma,p$.
 Using the above three estimates in Lemma \ref{Lemma:oracle} with $n'=c_{10}(N^*)^d$, $\mathcal U=N^*N^d/s$, we have
 \begin{eqnarray*}
      &&  Pr\{\|\pi_Mf_{ D,n,L}-f_\rho\|_\rho^2>\varepsilon+2\|h_\rho-f_\rho\|_\rho^2\}\nonumber\\
      &\leq&
      \exp\left\{c_{10}(N^*)^d\log\frac{16MN^dN^*}{s\varepsilon}-\frac{3m\varepsilon}{512M^2}\right\}\\
       &+&
      \exp\left\{\frac{-3m\varepsilon^2}{16(3M+C_5+1)^2\left(
    6\mathcal A_p+
    \varepsilon\right)}\right\}.
\end{eqnarray*}
Thus, by scaling $3\varepsilon$ to $\varepsilon$, for
$
 \varepsilon\geq \mathcal A_p,
$
we obtain
 \begin{eqnarray}\label{probability2}
      &&  Pr\{\|\pi_Mf_{ D,n,L}-f_\rho\|_\rho^2>\varepsilon\}\nonumber\\
      &\leq&
      \exp\left\{c_{10}(N^*)^d\log\frac{48MN^*N^d}{s\varepsilon}-\frac{m\varepsilon}{512M^2}\right\}\nonumber\\
       &+&
      \exp\left\{\frac{-m\varepsilon^2}{16(3M+C_5+1)^2\left(
    18\mathcal A_p+
    \varepsilon\right)}\right\}.
\end{eqnarray}
Thus,
\begin{eqnarray*}
      &&\mathbf E[\|\pi_Mf_{ D,n,L}-f_\rho\|_\rho^2]\\
     & =&\int_0^\infty{ Pr}[\|\pi_Mf_{ D,n,L}-f_\rho\|_\rho^2>\varepsilon]d\varepsilon\\
      &=&
      \int_{3\mathcal A_p}^\infty{ Pr}[\|\pi_Mf_{ D,n,L}-f_\rho\|_\rho^2>\varepsilon]d\varepsilon\\
      &+&
      \int_0^{3\mathcal A_p}{ Pr}[\|\pi_Mf_{ D,n,L}-f_\rho\|_\rho^2>\varepsilon]d\varepsilon\\
      &\leq&
       \int_{3\mathcal A_p}^\infty{ Pr}[\|\pi_Mf_{ D,n,L}-f_\rho\|_\rho^2>\varepsilon]d\varepsilon
      +
      3\mathcal A_p.
\end{eqnarray*}
From (\ref{probability2}), we also have
\begin{eqnarray*}
    &&\int_{3\mathcal A_p}^\infty{ Pr}[\|\pi_Mf_{ D,n,L}-f_\rho\|_\rho^2>\varepsilon]d\varepsilon\\
    &\leq&
    \int_{3\mathcal A_p}^\infty      \exp\left\{c_{10}(N^*)^d\log\frac{48MN^dN^*}{s\varepsilon}-\frac{m\varepsilon}{512M^2}\right\}d\varepsilon\\
      & +&
     \int_{3\mathcal A_p}^\infty  \exp\left\{\frac{-m\varepsilon^2}{16(3M+C_5+1)^2\left(
    18\mathcal A_p+
    \varepsilon\right)}\right\}  d\varepsilon\\
    &=:&
    J_1+J_2.
\end{eqnarray*}
A direct computation then yields
\begin{eqnarray*}
    J_2
    &\leq&
    \int_{3\mathcal A_p}^\infty  \exp\left\{\frac{-m\varepsilon}{112(3M+C_5+1)^2}\right\}  d\varepsilon\\
    &\leq&
    \frac{112(3M+C_5+1)^2}{m}.
\end{eqnarray*}
Set
\begin{equation}\label{def.N*}
      (N^*)^{2r+d}\sim
        c_{11}m\left(\frac{s}{N^d}\right)^\frac{2}p/\log(c_{11}^{1/2r+d}m) ,
\end{equation}
where  $
    c_{11}:=\frac{3C_4^2D_{\rho_X,p}^2}{c_{10}c_{11}1024M^2}$.
It follows from (\ref{restions-on-m-1})   that $N^*\geq\max\{\tilde{C},4N\}$. Thus, for  $p\geq 2$, we have, from the definition of $\mathcal A_p$, that
\begin{eqnarray*}
  && \log\frac{48MN^dN^*}{s\mathcal A_p}
   \leq
    \log\frac{48MN^{dp}(N^*)^{2r+1}}{C_4^2D^2_{\rho_X,p}s^{p}}\\
    &\leq&
    \log\frac{48M  (N^*)^{2r+1+dp}}{C_4^2D^2_{\rho_X,p}}
    \leq c_{12}\log N^*,
\end{eqnarray*}
where $c_{12}:=(2r+1+dp)\log\left(\frac{48M}{C_4^2D^2_{\rho_X,p}}+1\right)$.
 Then
$$
        c_{10}c_{12}(N^*)^d \log N^*\leq \frac{3m \mathcal A_p}{1024M^2},
$$
which implies
\begin{eqnarray*}
    &&
    J_1
    \leq
    \int_{3\mathcal A_p}^\infty  \exp\left\{\frac{-m\varepsilon}{1024M^2}\right\}  d\varepsilon
    \leq
    \frac{1024M^2}{m}.
\end{eqnarray*}
Thus,
$$
     \mathbf E[\|\pi_Mf_{ D,n,L}-f_\rho\|_\rho^2]
     \leq
     \frac{c_{13}}{m}+3\mathcal A_p,
$$
where $c_{13}:=1024M^2+112(3M+C_5+1)^2$. Hence,
\begin{eqnarray*}
   &&\mathbf E[\|\pi_Mf_{ D,n,L}-f_\rho\|_\rho^2]\\
     &\leq&
     C_7   \left(\frac{m}{\log m}\right)^\frac{-2r}{2r+d}  \left(\frac{s}{N^d}\right)^{\frac2p-\frac{2r}{2r+d}}.
\end{eqnarray*}
This provides the upper bound of (\ref{almost optimal learning rate}).
\end{IEEEproof}

\section{Proof of the Lower Bounds }\label{Sec.Lower bound}

In this section, we present a general lower bound estimate for Theorem \ref{Theorem:sampling-theorem} and Theorem \ref{Theorem: ERM}. To this end,
we need the following assumption for the qualification of the distribution $\rho$.

\begin{assumption}\label{Assumption on distribution} Assume

(A) $f_\rho\in Lip^{(N,s,r,c_0)}$.

(B) $\rho_X$ is the uniform distribution  on $\mathbb I^d$.

(C) $y=f_\rho(x)+\nu$, where $\nu$ and $x$ are independent and $\nu$
is drawn according to the standard normal distribution $\mathcal
N(0,1)$.
\end{assumption}

Let $\mathcal M_1(N,s,r,c_0)$ be the set of all distributions that satisfy Assumption \ref{Assumption on distribution} and $\Psi_m$ be the set of estimators $f_D$ derived from $D_m$. Then
\begin{eqnarray}\label{dis-comp}
   &&\sup_{\rho\in\mathcal
     M(N,s,r,c_0)}\inf_{f_D\in\Psi_m}\mathbf E[\|f_D-f_\rho\|_\rho^2]\nonumber\\
     &\geq&
     \sup_{\rho\in\mathcal
     M_1(N,s,r,c_0)}\inf_{f_D\in\Psi_m}\mathbf E[\|f_D-f_\rho\|_\rho^2].
\end{eqnarray}
The following theorem is a more general lower bound than that in Theorem \ref{Theorem:sampling-theorem}.

\begin{theorem}\label{Theorem:lower bound}
 If
$m$ satisfies (\ref{restions-on-m}), then there exists a constant
$\tilde{C}$ independent of $m$, $s$ or $N$, such that
\begin{equation}\label{theorem3}
     \sup_{\rho\in\mathcal
     M(N,s,r,c_0)}\inf_{f_D\in\Psi_m}\mathbf E[\|f_D-f_\rho\|_\rho^2]\geq
     \tilde{C}
     m^{-\frac{2r}{2r+d}}\left(\frac{s}{N^d}\right)^\frac{d}{2r+d}.
\end{equation}
\end{theorem}

It it easy to see that the lower bound of Theorem \ref{Theorem:sampling-theorem} is a direct consequence of Theorem \ref{Theorem:lower bound}. Before presenting the proof, we  introduce a function $g$ that satisfies
the following assumption.

\begin{assumption}\label{Assumption:g}
Assume that $g:\mathbb R^d\rightarrow\mathbb R$ satisfies
$supp(g)=[-/(2\sqrt{d}),1/(2\sqrt{d})]^d$, $g(x)=1$ for
$x\in[-1/(4\sqrt{d}),1/(4\sqrt{d})]^d$ and $g\in
Lip^{(r,c_02^{v-1})}$, where $supp(g)$ denotes the support of $g$.
\end{assumption}

Let $\{\epsilon_{k}\}_{k=1}^{(N^*)^d}$  be a set of independent
Rademacher random variables, i.e.,
\begin{equation}\label{Rademacher}
      Pr(\epsilon_{  k}=1)=Pr(\epsilon_{ k}=-1)=\frac12,\qquad
      \forall\ k=1,2,\dots,(N^*)^d.
\end{equation}
 For
$x\in\mathbb I^d$, define
\begin{equation}\label{Def.gk}
      g_k(x):=({N^*})^{-r}g({N^*}(x-\xi_k)),
\end{equation}
where $\xi_k$ is the center of the cube $B_k$.   Since
$$
          \| {N^*} (x-\xi_k)-{N^*} (x-\xi_{k'})\|
          = N^* \|\xi_k-\xi_{k'}\|\geq1,\qquad \forall\ k\neq k',
$$
at least one of
 $ {N^*}  (x-\xi_k)$ and $ {N^*} (x-\xi_{k'})$ lies outside
$(-1/(2\sqrt{d}),1/(2\sqrt{d}))^d$. Then it follows
from Assumption \ref{Assumption:g} that
\begin{equation}\label{gk property}
       g_k(x)=0,\qquad \mbox{if}\ x\notin \dot{B}_k,
\end{equation}
where $\dot{B}_k= B_k\backslash \partial B_k$ and $\partial A$ denotes the boundary of a cube $A$.

Given $S=\cup_{j\in\Lambda_s }A_{ j}$, consider the set $\mathcal F_{S,N^*}$ of all functions,
$$
     f(x)=\left\{\begin{array}{cc}
     \sum_{k=1}^{(N^*)^d}\epsilon_kg_k(x),& \mbox{if}\ x\in S, \nonumber\\
      0,&\mbox{otherwise,}\end{array}\right.
$$
with $\epsilon_k$  that satisfies  (\ref{Rademacher}).
It is obvious that $\mathcal F_{S,N^*}$ is a set of random
functions. The following lemma shows that it is almost surely a
subset of $ Lip^{(N,s,r,c_0)}$.

\begin{lemma}\label{Lemma:Subset}
If $g_k$ is defined by (\ref{Def.gk}) with $g$ satisfying the Assumption
\ref{Assumption:g}, then for  $N^*\in\mathbb N$ and
$S=\cup_{j\in\Lambda_s }A_{ j}$, then
$$
     \mathcal F_{S,N^*}\subset Lip^{(N,s,r,c_0)}
$$
almost surely.
\end{lemma}

\begin{IEEEproof} From the definition of $\mathcal F_{S,N^*}$, it is obvious that each $f\in
\mathcal F_{S,N^*}$ is $s$-sparse in $N^d$ partitions. So, it
suffices to prove that   $f\in \mathcal F_{S,N^*}$ implies $f\in
Lip^{(r,c_0)}$ almost surely.
 For
$x,x'\in \mathbb I^d$ with $x\neq x'$, we divide the proof into four
cases: $x,x'\in S$, $x\in S$ but $x'\notin S$, $x\notin S$ but
$x'\in S$ and $x,x'\not\in S$.

{\it Case 1:   $x,x'\in S$.} If $x,x'\in B_{k_0}\cap S$ for some
$k_0\in\{1,\dots,(N^*)^d\}$,
       then (\ref{gk property}) yields
        $(g_k)^{(u)}_{\alpha} (x)=0$ for $k\neq k_0$. So, for each  $f\in\mathcal F_{S,N^*}$, we  get from
$|\epsilon_k|=1$, (\ref{gk property}),  (\ref{Def.gk}) and  $0<v\leq
1$ that
\begin{eqnarray*}
      &&|f^{(u)}_{\alpha} (x)-f^{(u)}_{\alpha} (x')|\\
      &=&
      \left|\sum_{k=1}^{(N^*)^d}\epsilon_{k}(g_k)^{(u)}_{\alpha}(x)-(g_k)^{(u)}_{\alpha} (x')]\right|\\
       &=&
      \left| (g_k)^{(u)}_{\alpha}(x)-(g_k)^{(u)}_{0}( x') \right|\\
      &\leq& (N^*)^{-r+u}\left|g^{(u)}_{\alpha} ( N^* (x-\xi_{k_0}))-g^{(u)}_{\alpha} ( N^* (x'-\xi_{k_0}))\right|\\
       &\leq&
      c_0 2^{v-1}\|x-x'\|^v
      \leq
     c_0\|x-x'\|^v.
\end{eqnarray*}

If $x\in B_{k_1}\cap S$ but $x'\in B_{k_2}\cap S$ for some
$k_1,k_2\in\{1,\dots,(N^*)^d\}$  with $k_1\neq k_2$, we
      can
choose $z\in \partial B_{k_1}$ and $z'\in \partial B_{k_2}$ such
that $z,z'$ are on the segment between $x$ and ¡¡$x'$. Then
\begin{equation}\label{inverse triangle}
           \|x-z\|+\|x'-z'\|\leq \|x-x'\|.
\end{equation}
So,   Assumption \ref{Assumption:g}, (\ref{gk property}), $0<v\leq
1$, Jensen's inequality  and    (\ref{inverse triangle}) show
\begin{eqnarray*}
        &&|f^{(u)}_{\alpha}(x)-f^{(u)}_{\alpha}(x')|\\
      &=&
      \left|\sum_{k=1}^{(N^*)^d}\epsilon_{k}[(g_k)^{(u)}_{\alpha}(x)- (g_k)^{(u)}_{\alpha}(x')]\right|\\
       &\leq&
      \left|(g_{k_1})^{(u)}_{\alpha}(x)\right|+\left|(g_{k_2})^{(u)}_{\alpha} (x')\right|\\
      &=&
      \left|(g_{k_1})^{(u)}_{\alpha}(x)-(g_{k_1})^{(u)}_{\alpha} (z)\right|+\left|(g_{k_2})^{(u)}_{\alpha}(x')- (g_{k_2})^{(u)}_{\alpha} (z')
      \right|\\
       &\leq&
      (N^*)^{-r+u}\left[|g^{(u)}_{\alpha}( N^* (x-\xi_{k_1}))-g^{(u)}_{\alpha}( N^*
      (z-\xi_{k_1}))|\right.\\
        &+&
      \left.|g^{(u)}_{\alpha} (N^* (x'-\xi_{k_2}))-g^{(u)}_{\alpha}( N^* (z'-\xi_{k_2}))|\right]\\
      &\leq&
      c_02^v\left[\frac{\|x-z\|^v}2+\frac{\|x'-z'\|^v}2\right]\\
      &\leq&
       c_02^v\left[\frac{\|x-z\|}2+\frac{\|x'-z'\|}2\right]^v\leq
       c_0\|x-x'\|^v.
\end{eqnarray*}
These two  assertions imply that $f\in Lip^{(r,c_0)}$ almost surely and proves
Lemma \ref{Lemma:Subset} for the first case.

{\it Case 2: Suppose $x\in S$, $x'\notin S$.} There is some
$k_3\in\{1,\dots,(N^*)^d\}$ such that $x\in S \cap B_{k_3}$.
  For each $f\in \mathcal F_{S,N^*}$
and any $x'\notin S$, it follows from (\ref{Def.gk}) and
Assumption \ref{Assumption:g} that
$$
      f(x')=0=f(z), \qquad \forall\ z\in\partial B_{k_3}.
$$
Select a  $z''\in \partial B_{k_3}$ on the   segment between
$x$ and $x'$. Then,  $\|x-x'\|\geq \|x-z''\|$.
 Hence,  the result
in the first case above shows that
\begin{eqnarray*}
       &&|f^{(u)}_{\alpha}(x)-f^{(u)}_{\alpha}(x')|= |f^{(u)}_{\alpha}(x)-f^{(u)}_{\alpha}(z'')|\\
       &\leq& c_0\|x-z''\|^v\leq
       c_0\|x-x'\|^v.
\end{eqnarray*}

{\it Case 3: Suppose $x'\in S$, $x\notin S$.} The proof of this  case is the
same as that of Case 2.

{\it Case 4: Suppose $x,x'\notin S$.} For each $f\in\mathcal F_{S,N^*}$ and
any $x,x'\notin S$, we have
$$
      |f^{(u)}_{\alpha}(x)-f^{(u)}_{\alpha}(x')|=0\leq c_0\|x-x'\|^v.
$$
Combining the above four cases, we complete the proof of Lemma
\ref{Lemma:Subset}.
\end{IEEEproof}

Let  $\mathcal
   H_{S,N^*}$ be the set of all functions
$$
  h(x)=
    \left\{\begin{array}{cc}
     \sum_{k=1}^{(N^*)^d}c_kg_k(x),&  \mbox{if}\ x\in S,\\
      0,&\mbox{otherwise}\end{array}\right.
$$
with $c_k\in \mathbb
   R $.
It then follows  from the definition of $\mathcal F_{S,N^*}$ that
\begin{equation}\label{subsubset}\
       \mathcal F_{S,N^*}\subset \mathcal
       H_{S,N^*}.
\end{equation}
  The following
lemma constructs an orthonormal basis of $\mathcal H_{S,N^*}$.

\begin{lemma}\label{Lemma:orthonormal basis}
Let $\mathcal H_{S,N^*}$ be defined as above with $g_k$
  and $g$ that satisfy (\ref{Def.gk})  and Assumption
\ref{Assumption:g}, respectively.  Let
\begin{equation}\label{Def g*}
    g^*_{k,S}(x):=\left\{\begin{array}{cc} g_{k}(x),& \mbox{if}\ x\in S,\\ 0,&  \mbox{if}\ x\notin
      S.\end{array}\right.
\end{equation}
Then, the system $
       \left\{\frac{g^*_{k,S}(\cdot)}{\| g^*_{  k,S}\|_\rho}: k=1,\dots,(N^*)^d\right\}
$ is an orthonormal basis of   $\mathcal H_{S,N^*}$ using the inner product
of $L_{\rho_X}^2$.
\end{lemma}

\begin{IEEEproof}
 For   $k\neq k'$, it follows from (\ref{gk property}) and (\ref{Def g*})
  that
\begin{eqnarray}\label{orthogonality}
          \int_{\mathbb I^d}g^*_{  k,S}(x)g^*_{  k',S}(x)d\rho_X
         =\int_{S}g_{  k}(x)g_{  k'}(x)d\rho_X
      =0.
\end{eqnarray}
 Therefore, $\{g_{k,S}^*(\cdot): k=1,\dots,(N^*)^d\}$ is an orthogonal set in  $L_{\rho_X}^2$.
 Noting further  $\|g_{k,S}^*\|_\rho\neq 0$ for all $k\in
\{1,\dots,(N^*)^d\}$,
$$
      \int_{\mathbb I^d}\left(\frac{g^*_{  k,S}(
        x)}{\|g_{k,S}^*\|_\rho}\right)^2d\rho_X=1,\qquad \forall\
        k=1,2,\dots,(N^*)^d
$$
and  $\mathcal H_{S,N^*}$ is  an $(N^*)^d$-dimensional linear space,
we may conclude  that the system $
       \left\{\frac{g^*_{k,S}(\cdot)}{\| g^*_{  k,S}\|_\rho}: k=1,\dots,(N^*)^d\right\}
$ is an orthonormal basis of   $\mathcal H_{S,N^*}$. This
  completes the proof of Lemma
\ref{Lemma:orthonormal basis}.
\end{IEEEproof}

To prove the lower bound, we need the following three lemmas. The
first one can be found in \cite[Lemma 3.2]{Gyorfi2002}.

\begin{lemma}\label{Lemma:concentration}
Let $U$ be an $\ell$-dimensional real vector, $\theta$  a zero-
mean random variable with range $\{-1,1\}$, and $\nu$  an
$\ell$-dimensional random vector of standard normal variable, independent of
$U$. Denote
$$
         \psi:=\theta U+\nu.
$$
Then there exists an absolute constant $\tilde{C}_1>0$ such that
$$
      \min_{f^*:\mathbb R^{\ell}\rightarrow \{-1,1\}}Pr\{f^*(\psi)\neq
      \theta\}\geq \tilde{C}_1e^{-\|U\|_{\ell}^2/2},
$$
where $\|\cdot\|_{\ell}$ denotes the $\ell$-dimensional Euclidean
norm and the minimization is over all functions $f^*:\mathbb
R^{\ell}\rightarrow  \{-1,1\}$.
\end{lemma}

\begin{lemma}\label{Lemma:lower bound for integral}
Under (B) in Assumption \ref{Assumption on distribution}, if $g$
satisfies Assumption \ref{Assumption:g}, then for any $k\in \{1,\dots,(N^*)^d\}$
\begin{equation}\label{lower case 1}
     \int_{B_k}\left[g(N^*(x-\xi_k))
       \right]^2d\rho_X
      \geq  \tilde{C}_2(N^*)^{-d},
\end{equation}
where the constant $\tilde{C}_2$ dependent only on  $d$.
\end{lemma}

\begin{IEEEproof}
It  follows from Assumption \ref{Assumption:g} and (B) that
\begin{eqnarray}\label{norm equv}
        &&\int_{B_k}\left[g(N^*(x-\xi_k))
       \right]^2d\rho_X\nonumber\\
      &\geq&
      \int_{ B_k}\left[g(N^*(x-\xi_k))
       \right]^2dx \nonumber\\
      &\geq&
      (N^*)^{-d}\int_{[-1/(2\sqrt{d}),1/(2\sqrt{d})]^d}|g(x)
       |^2dx\nonumber\\
       &\geq&
       (N^*)^{-d}\int_{[-1/(4\sqrt{d}),1/(4\sqrt{d})]^d}dx\nonumber\\
      &= &
      (2\sqrt{d}N^*)^{-d},
\end{eqnarray}
where the second inequality holds since $N^*(x-\xi_k)$ is restricted to some subset of
$\mathbb R^d$ that contains $[-1/(2\sqrt{d}),1/(2\sqrt{d})]^d$ for $x \in B_k$.
 This completes the
proof of Lemma \ref{Lemma:lower bound for integral} with
$\tilde{C}_2=\tilde{C}_3(2\sqrt{d})^{-d}$.
\end{IEEEproof}

If $N^*\geq 4N$, noting that
   $\{A_j\}_{j=1}^{N^d}$ and
$\{B_k\}_{k=1}^{(N^*)^d}$ are cubic partitions of $\mathbb I^d$, we may conclude that
each $A_j$ then contains at least
$\left(\frac{N^*}{N}-2\right)^d\geq \left(\frac{N^*}{2N}\right)^d$
$B_{k}$'s. For each $j\in\Lambda_s$,  denote
\begin{equation}\label{Lambda*}
   \Lambda^*_{j }:=\{k\in\{1,\dots,(N^*)^d\}: B_k\subseteq A_j\}.
\end{equation}
Then
\begin{equation}\label{card lambda *}
     |\Lambda^*_{j}|\geq
       \left(\frac{N^*}{2N}\right)^d.
\end{equation}
With the above preparations , we present the following  lemma, which will play a
crucial role  in our analysis.

\begin{lemma}\label{Lemma: our concentration}
Let
$D_m=\{(x_i,y_i)\}_{i=1}^m$ be the set of samples which are
independently  drawn according to some distribution
$\rho$ with the marginal distribution $\rho_X$ satisfying (B) and
$y_i=f_\rho(x_i)+\nu_i$, where $f_\rho\in\mathcal F_{S,N^*}$ and
$\nu_i$ is the standard normal variable. If  $N^*\geq 4N$ and
$N^*=\left\lceil
\left(\frac{ms}{N^d}\right)^{\frac{1}{2r+d}}\right\rceil$. Then for any $j\in\Lambda_s$, $k\in\Lambda_j^*$,
there exists a constant $\tilde{C}_3$ independent of $m$, $s$, $N$
or $N^*$ such that
\begin{equation}\label{lower prob case1}
      \min_{h:\mathbb R^{m}\rightarrow\{-1,1\}}Pr\{h((y_1,\dots,y_m))\neq
      \epsilon_{ k}\}\geq
      \tilde{C}_3>0.
\end{equation}
\end{lemma}

\begin{IEEEproof}
Write $D_{in}=\{x_i\}_{i=1}^m$ and
$D_{in,S}:=D_{in}\cap S$. For each $j\in\Lambda_s$ and $k\in \Lambda_j^*$, denote  further $ B_{k,D}:=B_k\cap D_{in}:=\{
x_{i,{ k}}\}_{i=1}^{\ell'}$, where $\ell'=0$ means
$B_{k,D}=\varnothing$. We  then divide the proof into the following
three steps.

{\it Step 1: Estimating $|D_{in,S}|$.} Since $\rho_X$ is the uniform
distribution on $\mathbb I^d$, for each $x_i\in D_{in}$,
$$
      Pr\{x_i\in S\}=\frac{s}{N^d}.
$$
For $i=1,\dots,m$, define
$$
         V_i:=\mathcal I_{x_i\in S}:=\left\{\begin{array}{cc}
         1,& \mbox{with probability}\  \frac{s}{N^d}\\
         0,& \mbox{with probability}
         \ 1-\frac{s}{N^d}.\end{array}\right.
$$
Then
$$
      |D_{in,S}|=\sum_{i=1}^mV_i=\sum_{i=1}^m\mathcal I_{x_i\in S}.
$$
This implies
\begin{eqnarray*}
     &&\mathbf
     E\{|D_{in,S}|\}=\sum_{i=1}^m\mathbf E\{\mathcal I_{x_i\in
     S}\}=\sum_{i=1}^mPr\{x_i\in S\}\\
     &=&
     \sum_{i=1}^m\frac{s}{N^d}=\frac{ms}{N^d}.
\end{eqnarray*}
So,  it follows from Markov's inequality that
\begin{equation*}\label{number 1}
       Pr\left\{|D_{in,S}|>\left\lceil\frac{2ms}{N^d}\right\rceil\right\}\leq \frac{N^d\mathbf
     E\{|D_{in,S}|\}}{ 2ms}=\frac12.
\end{equation*}
The above estimate, together with  the formula of total probability,
implies that
\begin{eqnarray}\label{lower 1.1}
     &&\min_{h:\mathbb R^{m}\rightarrow\{-1,1\}}Pr\{h(y_D)\neq
      \epsilon_{ k} \}  \\
      &=&
      \min_{h:\mathbb R^{m}\rightarrow\{-1,1\}}Pr\left\{h(y_D)\neq
      \epsilon_{ k}\big| |D_{in,S}|>\left\lceil\frac{2ms}{N^d}\right\rceil\right\}\nonumber\\
      &&Pr\left\{|D_{in,S}|>\left\lceil\frac{2ms}{N^d}\right\rceil\right\}\nonumber\\
     &+&
      \min_{h:\mathbb R^{m}\rightarrow\{-1,1\}}Pr\left\{h(y_D)\neq
      \epsilon_{ k}\big| |D_{in,S}|\leq\left\lceil\frac{2ms}{N^d}\right\rceil\right\}\nonumber\\
      &&Pr\left\{|D_{in,S}|\leq\left\lceil\frac{2ms}{N^d}\right\rceil\right\}\nonumber\\
      &\geq&
      \frac12\min_{h:\mathbb R^{m}\rightarrow\{-1,1\}}Pr\left\{h(y_D)\neq
      \epsilon_{ k}\big| |D_{in,S}|\leq\left\lceil\frac{2ms}{N^d}\right\rceil\right\},\nonumber
\end{eqnarray}
where $h(y_D):=h((y_1,\dots,y_m))$.

{\it Step 2: Estimating the conditional probability.} If $\mathcal
A$ and $\mathcal B$ are random events, then
\begin{equation}\label{Probability to expectation}
        Pr\{\mathcal A\}=\mathbf E\{\mathcal I_{\mathcal A}\}=
        \mathbf E\left\{\mathbf E\{\mathcal I_{\mathcal A}|\mathcal
        B\}\right\}=\mathbf E\left\{Pr\{\mathcal A|\mathcal
        B\}\right\},
\end{equation}
where $\mathcal I_{\mathcal A}$ denotes the indicator of the event
$\mathcal A$. Hence,
\begin{eqnarray}\label{lower 1.2}
      &&\min_{h:\mathbb R^{m}\rightarrow\{-1,1\}}Pr\left\{h(y_D)\neq
      \epsilon_{ k}\big| |D_{in,S}|\leq\left\lceil\frac{2ms}{N^d}\right\rceil\right\} \nonumber\\
      &=&
       \mathbf E\big\{\min_{h:\mathbb R^{m}\rightarrow\{-1,1\}}Pr\{h(y_D)\neq
      \epsilon_{
      k}\big|\nonumber\\
      &&|D_{in,S}|\leq\left\lceil 2ms/N^d \right\rceil,D_{in}\}\big\}.
\end{eqnarray}
For each $j\in\Lambda_s$ and $k\in \Lambda_j^*$, it follows from
(\ref{Lambda*}) that $\ell'=|B_{k,D}|\leq |D_{in,S}|$. Then for
 each $h:\mathbb
R^{m}\rightarrow\{-1,1\}$, from the  formula of total probability
again, we obtain
\begin{eqnarray}\label{lower 1.3}
     &&Pr\left\{h(y_D)\neq
      \epsilon_{
      k}\big||D_{in,S}|\leq\left\lceil\frac{2ms}{N^d}\right\rceil,D_{in}\right\}
      \\
      &=&
      \sum_{\ell=0}^{\left\lceil\frac{2ms}{N^d}\right\rceil}Pr\big\{h(y_D)\neq
      \epsilon_{
      k}\big||D_{in,S}|\leq\left\lceil\frac{2ms}{N^d}\right\rceil,\nonumber\\
      &&D_{in},\ell'=\ell\big\}Pr\left\{\ell'=\ell\big||D_{in,S}|\leq\left\lceil\frac{2ms}{N^d}\right\rceil,D_{in}\right\}\nonumber
\end{eqnarray}
and
\begin{equation}\label{lower 1.4}
      \sum_{\ell=0}^{\left\lceil\frac{2ms}{N^d}\right\rceil}Pr\left\{\ell'=\ell\big||D_{in,S}|\leq\left\lceil\frac{2ms}{N^d}\right\rceil,D_{in}\right\}
      =1.
\end{equation}

Given $D_{in}$, $\ell'=0$,
$|D_{in,S}|\leq\left\lceil\frac{2ms}{N^d}\right\rceil$, for each
$k\in\{1,\dots,(N^*)^d\}$,  it follows from the definition of $\mathcal F_{S,N^*}$  and
(\ref{gk property})
  that there exists some ${  k}'\neq {  k}$ such that
$$
       y_i=\sum_{k=1}^{(N^*)^d}
      \epsilon_{ k}g_{ k}(x_i)+\nu_i= \epsilon_{{ k}'}g_{{
      k}'}(x_i)+\nu_i, \qquad i=1,2,\dots,m,
$$
which is  independent of $\epsilon_{ k}$. That is, $\epsilon_k$ is
independent of $(y_1,\dots,y_m)$. Thus, it follows from (\ref{Rademacher}) that
\begin{eqnarray}\label{probbability 2}
       &&\min_{h:\mathbb R^{m}\rightarrow\{-1,1\}}
      Pr\big\{h((y_1,\dots,y_m)\neq
      \epsilon_{ k}\big| \nonumber\\
      &&
      |D_{in,S}|\leq\left\lceil 2ms/N^d \right\rceil,D_{in},\ell'=0\big\}=\frac12.
\end{eqnarray}

Given $D_{in}$,
$|D_{in,S}|\leq\left\lceil\frac{2ms}{N^d}\right\rceil$ and
$\ell'=\ell$ with $\ell\geq 1$, for each $j\in\Lambda_s$ and $k\in \Lambda_j^*$,
  then   we get from
the definition of $\mathcal F_{S,N^*}$  and (\ref{gk property})
  that there exists a ${  k}'\neq {  k}$, such that
$$
       y_i=\sum_{k=1}^{(N^*)^d}
      \epsilon_{ k}g_{ k}(x_i)+\nu_i= \epsilon_{{ k}'}g_{{
      k}'}(x_i)+\nu_i, \qquad x_i\in  D_{in}\backslash B_{k,D}
$$
which is  independent of $\epsilon_{ k}$. Write
$$
       y_{i,k}=\sum_{k=1}^{(N^*)^d}
      \epsilon_{ k}g_{ k}(x_{i,k})+\nu_i=\epsilon_{ k}g_{ k}(x_{i,k})+\nu_i,  \qquad i=1,\dots,\ell'.
$$
Then, there exists an $h^*:\mathbb R^{\ell'}\rightarrow\{-1,1\}$, such that
\begin{eqnarray}\label{insert 1}
       &&
      Pr\left\{h(y_D)\neq
      \epsilon_{ k}\big||D_{in,S}|\leq\left\lceil\frac{2ms}{N^d}\right\rceil,D_{in},\ell'=\ell \right\} \\
     & =&
       Pr\left\{h^*(y_{D,\ell'})\neq
      \epsilon_{ k}\big||D_{in,S}|\leq\left\lceil\frac{2ms}{N^d}\right\rceil,D_{in},\ell'=\ell\right\}, \nonumber
\end{eqnarray}
where $y_{D,\ell'}:=(y_{1,{ k}},\dots,y_{\ell',{ k}})$.
From (\ref{gk property}) again, it is easy to see that
\begin{eqnarray}\label{y part}
         &&\left(y_{1,{  k}},\dots,y_{\ell',{  k}}\right) \\
         &:=&
         \epsilon_{ k}\left(g_{k}(x_{1,{k}}),\dots,g_{ k}(x_{\ell',{
         k}})\right)+(\nu_{1,{ k}},\dots,\nu_{\ell',{ k}}),\nonumber
\end{eqnarray}
Therefore,
   applying  Lemma
\ref{Lemma:concentration} with $U=\left(g_k(x_{1,{
k}}),\dots,g_k(x_{\ell',{ k}})\right)$ and $\theta=\epsilon_{ k}$,
we  get from (\ref{y part}) and (\ref{insert 1}) that for each $k\in
\Lambda_j^*$ and $j\in\Lambda_s$,
\begin{eqnarray}\label{lower 1.5}
      &&\min_{h^*:\mathbb R^{\ell'}\rightarrow\{-1,1\}}
      Pr\left\{h^*(y_{D,\ell'})\neq
      \epsilon_{ k}\big||D_{in,S}|\leq\big\lceil\frac{2ms}{N^d}\right\rceil,\nonumber\\
      &&D_{in},\ell'=\ell\big\} \nonumber\\
      &\geq& \tilde{C}_1\exp\left(-\frac{(g_{k}(x_{1,{ k}}))^2+
      \dots+(g_{k}(x_{\ell,{
      k}}))^2}2\right).
\end{eqnarray}
Putting (\ref{lower 1.5}) and (\ref{probbability 2}) into
(\ref{lower 1.3}) and noting (\ref{insert 1}) and (\ref{lower 1.4}),
we obtain that for each $k\in \Lambda_j^*$ and $j\in\Lambda_s$,
\begin{eqnarray}\label{lower 1.6}
     &&\min_{h:\mathbb R^{m}\rightarrow\{-1,1\}}Pr\big\{h(y_D)\neq
      \epsilon_{
      k}\big||D_{in,S}|\leq\left\lceil\frac{2ms}{N^d}\right\rceil,\nonumber\\
      &&D_{in}\big\}
       \geq
      \frac12
      Pr\left\{\ell'=0\big||D_{in,S}|\leq\left\lceil\frac{2ms}{N^d}\right\rceil,D_{in}\right\}
      \nonumber
      \\
      &+&
      \tilde{C}_1\sum_{\ell=1}^{\left\lceil\frac{2ms}{N^d}\right\rceil}\exp\left(-\frac{(g_{k}(x_{1,{ k}}))^2+
      \dots+(g_{k}(x_{\ell,{
      k}}))^2}2\right) \nonumber\\
      &&Pr\left\{\ell'=\ell\big||D_{in,S}|\leq\left\lceil\frac{2ms}{N^d}\right\rceil,D_{in}\right\}
      \nonumber\\
      &\geq&
      \min\left\{\frac12,\tilde{C}_1\mathcal B(m,s,N,g_k)\right\}.
\end{eqnarray}
where
$$
   \mathcal B(m,s,N,g_k):=\exp\left(-\frac{\sum_{x_i\in D_{in,S},|D_{in,S}|\leq \left\lceil\frac{2ms}{N^d}\right\rceil}
      (g_{k}(x_{i}))^2}2\right).
$$

{\it Step 3: Estimating the probability.} Putting  (\ref{lower 1.6})
into (\ref{lower 1.2}), we have, from Jensen's inequality with the
convexity of $\exp(-\cdot)$, that
\begin{eqnarray*}
      &&\min_{h:\mathbb R^{m}\rightarrow\{-1,1\}}Pr\left\{h(y_D)\neq
      \epsilon_{ k}\big| |D_{in,S}|\leq\left\lceil\frac{2ms}{N^d}\right\rceil\right\} \nonumber\\
      &\geq&
       \mathbf E\left\{\min\left\{\frac12,\tilde{C}_1\mathcal B(m,s,N,g_k)\right\}\right\}\\
      &\geq&
      \min\left\{\frac12,\tilde{C}_1\mathcal C(m,s,N,g_k)\right\},
\end{eqnarray*}
where
\begin{eqnarray*}
   &&\mathcal C(m,s,N,g_k)\\
   &:=&
   \exp\left(-\frac{\mathbf
      E\left\{\sum_{x_i\in D_{in,S},|D_{in,S}|\leq \left\lceil\frac{2ms}{N^d}\right\rceil}(g_{k}(x_{i}))^2\right\}}2\right).
\end{eqnarray*}
But (\ref{lower case 1}) implies that for each $j\in\Lambda_s$ and $k\in\Lambda_j^*$,
\begin{eqnarray}\label{insert 11111}
     &&\int_{\mathbb I^d} g^2_{k}( x)d\rho_X
        =
        \int_{ B_k} g^2_{k}(x)d\rho_X\\
        &=&
        (N^*)^{-2r}
        \int_{B_k}\left[g(N^*(x-\xi_k)) \right]^2d\rho_X
        \geq
        \tilde{C}_2(N^*)^{-2r-d}, \nonumber
\end{eqnarray}
which yields
$$
   \mathbf E\left\{\sum_{x_i\in
      D_{in,S},|D_{in,S}|\leq\left\lceil\frac{2ms}{N^d}\right\rceil}(g_{k}(x_{i}))^2 \right\}
      \geq
     \tilde{C}_2(N^*)^{-2r-d} \frac{2ms}{N^d}.
$$
Therefore, for each $j\in\Lambda_s$ and $k\in\Lambda_j^*$
\begin{eqnarray*}
       &&\min_{h:\mathbb R^{m}\rightarrow\{-1,1\}}Pr\left\{h(y_D)\neq
      \epsilon_{ k}\big| |D_{in,S}|\leq\left\lceil\frac{2ms}{N^d}\right\rceil\right\}\\
      &\geq&
      \min\left\{\frac12,\tilde{C}_1 \exp\left(-\frac{\tilde{C}_2(N^*)^{-2r-d} \frac{2ms}{N^d}
      }2\right)\right\}.
\end{eqnarray*}
Inserting the above estimate into (\ref{lower 1.1}), we then have
\begin{eqnarray*}
       &&\min_{h:\mathbb R^{m}\rightarrow\{-1,1\}}Pr\left\{h(y_D)\neq
      \epsilon_{ k}\right\}\\
      &\geq&
      \frac12\min\left\{\frac12,\tilde{C}_1 \exp\left(-\frac{\tilde{C}_2(N^*)^{-2r-d} \frac{2ms}{N^d}
      }2\right)\right\}.
\end{eqnarray*}
Since $N^*=\left\lceil
\left(\frac{ms}{N^d}\right)^{1/(2r+d)}\right\rceil$, we see that, for any $k\in\Lambda_j^*,j\in\Lambda_s$,
$$
 \min_{h:\mathbb R^{m}\rightarrow\{-1,1\}}Pr\{h((y_1,\dots,y_m))\neq
      \epsilon_{ k}\}\geq \tilde{C}_3
$$
with
  $\tilde{C}_3=\frac12\min\{1/2,\tilde{C}_1e^{-\tilde{C_2}/2}\}$.
 This
completes the proof of Lemma \ref{Lemma: our concentration}.
\end{IEEEproof}

We are now in a position to prove our main result.

{\bf Proof of Theorem \ref{Theorem:lower bound}.} For
$f_D\in\Psi_m$, define
\begin{eqnarray}\label{hat f}
       \hat{f}_D(x) &:=&
       \sum_{k=1}^{(N^*)^d}
       \frac{\int_{\mathbb I^d}f_D(x)g^*_{  k,S}(x)d\rho_X}{\| g^*_{  k,S}\|_\rho}
       g^*_{  k,S}(x)\nonumber\\
       &=:&\sum_{k=1}^{(N^*)^d}
       \hat{\epsilon}_{ k}g^*_{k,S}(x),
\end{eqnarray}
where $g^*_{k,S}$ is defined by (\ref{Def g*}). In view of  Lemma
\ref{Lemma:orthonormal basis}, we observe that  $\hat{f}_D$ is the orthogonal
projection of $f_D$ to $\mathcal H_{S,N^*}$. For $N^*\geq 4N$ and
$f_\rho^\epsilon\in\mathcal F_{S,N^*}\subset \mathcal H_{S,N^*}$
with $\epsilon=(\epsilon_1,\dots,\epsilon_{(N^*)^d})$ and
$\epsilon_k$ the Rademacher random variable,  it then follows from
(\ref{subsubset}), (\ref{Def g*}), (\ref{gk property}) and
(\ref{Lambda*})
  that
\begin{eqnarray*}
      &&\|f_D-f_\rho^\varepsilon\|_\rho^2\geq\|\hat{f}_D-f_\rho^\varepsilon\|_\rho^2\\
      &\geq&\sum_{j\in\Lambda_s}\sum_{k'\in\Lambda^*_{j }}
  \int_{B_{k'}}
   [\hat{f}_D(x)-f_\rho^\varepsilon(x)]^2d\rho_X\\
   &=&  \sum_{j\in\Lambda_s}\sum_{k'\in\Lambda^*_{j }}
  \int_{B_{k'}}
   \left[\sum_{k=1}^{(N^*)^d} (\hat{\epsilon}_{ k}-\epsilon_{k})g_{ k}(x)\right]^2d\rho_X\\
   &=&
   \sum_{j\in\Lambda_s}\sum_{k'\in\Lambda^*_{j }}
  \int_{B_{k'}}
   [\hat{\epsilon}_{ k'}-\epsilon_{k'}]^2[g_{ k'}(x)]^2d\rho_X\\
     & =&
      (N^*)^{-2r} \sum_{j\in\Lambda_s}\sum_{k'\in\Lambda^*_{j }}[\hat{\epsilon}_{ k'}
      -\epsilon_{ k'}]^2\\
      &\times&\int_{B_{k'}}\left[g(N^*(x-\xi_{k'}))\right]^2d\rho_X.
\end{eqnarray*}
 Define
$\tilde{\epsilon}_{k}=\left\{\begin{array}{cc} 1,& \hat{\epsilon}_{
k}\geq 0\\ -1, & \hat{\epsilon}_{ k}<0.\end{array}\right.$ Noting
that $\tilde{\epsilon}_{k}$ is a decision of $\epsilon_k$ based on
$D$, we may conclude that there exists some $h_k:\mathbb R^m\rightarrow\{-1,1\}$ such
that $h_k(y_1,\dots,y_m)=\tilde{\epsilon}_{k}$. Since $
      |\hat{\epsilon}_{ k}-\epsilon_{k}|
      \geq\frac{|\tilde{\epsilon}_{ k}-\epsilon_{k}|}2,
$ we have from    Lemma \ref{Lemma:lower bound for integral} that
\begin{eqnarray*}
     &&\|f_D-f_\rho^\varepsilon\|_\rho^2
     \geq
      \frac{\tilde{C}_2}4(N^*)^{-2r-d} \sum_{j\in\Lambda_s}\sum_{k'\in\Lambda^*_{j }}[\tilde{\epsilon}_{ k'}
      -\epsilon_{ k'}]^2\\
      &\geq& \frac{\tilde{C}_2}4(N^*)^{-2r-d} \sum_{j\in\Lambda_s}\sum_{k'\in\Lambda^*_{j }}\mathcal I_{ \tilde{\epsilon}_{ k'}
       \neq\epsilon_{k'} }.
\end{eqnarray*}
Hence,
\begin{eqnarray*}
      &&\inf_{f_D\in\Psi_m} \mathbf E[\|f_D-f_\rho^\varepsilon\|_\rho^2]\\
      &\geq&
      \frac{\tilde{C}_2}4(N^*)^{-2r-d} \inf_{\tilde{\epsilon}=(\tilde{\epsilon}_1,\dots,\tilde{\epsilon}_{(N^*)^d})}\sum_{j\in\Lambda_s}\sum_{k'\in\Lambda^*_{j
      }}Pr\{\tilde{\epsilon}_{k'}
       \neq\epsilon_{ k'}\}.
\end{eqnarray*}
But Lemma  \ref{Lemma: our concentration} and (\ref{card lambda *})
assure that for any set of independent Rademacher random
variables $\epsilon=(\epsilon_1,\dots,\epsilon_{(N^*)^d})$,
\begin{eqnarray*}
       &&\inf_{\tilde{\epsilon}}\sum_{j\in\Lambda_s}\sum_{k'\in\Lambda^*_{j
      }}Pr\{\tilde{\epsilon}_{k'}
       \neq\epsilon_{ k'}\}\\
       &=&\sum_{j\in\Lambda_s}\sum_{k'\in\Lambda^*_{j
      }}   \inf_{\tilde{\epsilon}_{k'}} Pr\{\tilde{\epsilon}_{k'}
       \neq\epsilon_{ k'}\}
        \geq \tilde{C}_3s\left(\frac{N^*}{2N}\right)^d.
\end{eqnarray*}
Therefore,  Lemma \ref{Lemma:Subset}, together with (\ref{dis-comp}), yields
\begin{eqnarray*}
     &&\sup_{\rho\in\mathcal
     M(N,s,r)}\inf_{f_D\in\Psi_m}\mathbb E[\|f_D-f_\rho\|_\rho^2]\\
     &\geq&
      \sup_{\epsilon}\inf_{f_D\in\Psi_m} \mathbf E[\|f_D-f_\rho^\varepsilon\|_\rho^2]\\
      &\geq&
      \frac{\tilde{C}_2}4(N^*)^{-2r-d}\tilde{C}_3s\left(\frac{N^*}{2N}\right)^d
       =
        \frac{\tilde{C}_2\tilde{C}_3}{2^{d+2}}(N^*)^{-2r}\frac{s}{N^d}.
\end{eqnarray*}
By setting $N^*=\left\lceil
\left(\frac{ms}{N^d}\right)^{1/(2r+d)}\right\rceil$, (\ref{restions-on-m})
implies $N^*\geq 4N$. Hence,,
$$
   \sup_{\rho\in\mathcal
     M(N,s,r,c_0)}\inf_{f_D\in\Psi_m}\mathbb E[\|f_D-f_\rho\|_\rho^2]
     \geq
     \tilde{C}
     m^{-\frac{2r}{2r+d}}\left(\frac{s}{N^d}\right)^{\frac{d}{2r+d}},
$$
where $\tilde{C}:=\frac{\tilde{C}_2\tilde{C}_3}{2^{d+2}}$.  This
completes the proof of Theorem \ref{Theorem:lower bound}. $\Box$

\begin{IEEEproof}[Proof of Theorem \ref{Theorem: ERM}] The upper bound of (\ref{almost optimal learning rate}) was established in Section \ref{Sec.Upperbound}. The lower bound of (\ref{almost optimal learning rate}) is a direct corollary of Theorem \ref{Theorem:lower bound}. This completes the proof of Theorem \ref{Theorem: ERM}.
\end{IEEEproof}

\begin{IEEEproof}[Proof of Theorem \ref{Theorem:sampling-theorem}]
The upper bound of (\ref{samplingtheorem}) can be derived from (\ref{almost optimal learning rate}) with $p=2$ and the lower bound is a consequence of Theorem \ref{Theorem:lower bound}. This completes the proof of Theorem \ref{Theorem:sampling-theorem}.
\end{IEEEproof}

\section*{Acknowledgement}
The research of CKC and BZ were partially supported by Hong
Kong Research Council [Grant Nos. 12300917 and 12303218] and Hong
Kong Baptist University [Grant Nos. RC-ICRS/16-17/03 and RC-FNRA-IG/18-19/SCI/01]. The
research of SBL was supported by the National Natural Science
Foundation of China [Grant Nos. 61876133,11771012], and the research of DXZ
was partially supported by the Research Grant Council of Hong Kong [Project No. {  CityU 11306318}] and carried out during his visit to the Erwin-Schroedinger Institute in   August, 2019.


\begin{thebibliography}{99}

\bibitem{Akkus2017}
Z. Akkus, A.  Galimzianova, A. Hoogi, D. L. Rubin, and B. J. Erickson. Deep learning for brain MRI segmentation: state of the art and future directions. J. Digital Imag., 30(4): 449-459, 2017.

%
%
%
\bibitem{Chherawala2016}
Y. Chherawala, P. P. Roy, and M. Cheriet. Feature set evaluation for offline handwriting recognition systems: application to the recurrent neural network model. IEEE Trans. Cyber., 46(12): 2825-2836, 2016.

 \bibitem{Chui1994}
C. K. Chui, X. Li, and H. N. Mhaskar. Neural networks for localized
approximation. Math. Comput., 63:  607-623, 1994.


\bibitem{Chui2018}
C. K. Chui, S. B. Lin, and D. X. Zhou. Construction of neural networks




for realization of localized deep
  learning.  Front. Appl. Math. Statis., 4: 14, 2018.

\bibitem{Chui2018a}
C. K. Chui, S. B. Lin, and D. X. Zhou. Deep neural networks for rotation-invariance approximation and learning. Anal. Appl.,   17: 737-772, 2019.

\bibitem{Chui2019}
 C. K. Chui, S. B. Lin, and D. X.  Zhou. Deep net tree structure for balance of capacity and approximation ability. Front. Appl. Math.   Statis.,   5: 46, 2019.


\bibitem{Ciresan2010}
D. C. Ciresan, U. Meier, L. M. Gambardella, and J. Schmidhuber. Deep, big, simple neural nets for handwritten digit recognition. Neural Comput., 22(12): 3207-3220, 2010.



\bibitem{Cucker2007}
F. Cucker and D. X. Zhou. Learning Theory: An Approximation Theory
Viewpoint.  Cambridge University Press, Cambridge, 2007.


\bibitem{Deluna2005}
X. De Luna and M. G. Genton. Predictive spatio-temporal models for spatially sparse enviromental data. Statis. Sinica, 15: 547-568, 2005.
%

\bibitem{Donoho2006}
D. L. Donoho. Compressed sensing. IEEE Trans. Inf. Theory,  52: 1289-1306, 2006.

\bibitem{ELayach2014}
O. El Ayach, S. Rajagopal, S. Abu-Surra, Z. Pi, and R. W. Heath. Spatially sparse precoding in millimeter wave MIMO systems. IEEE Trans. Wireless Commun., 13: 1499-1513, 2014.


\bibitem{Goodfellow2016}
I. Goodfellow, Y. Bengio, and A. Courville. {Deep Learning}. MIT
Press, 2016.

\bibitem{Graham2014}
B. Graham. Spatially-sparse convolutional neural networks. arXiv preprint arXiv:1409.6070.


\bibitem{Gittens2016}
A. Gittens and M. W. Mahoney.  Revisitng the Nystr\"{o}m method for
improved large scale machine learning.  J.  Mach. Learn. Res.,
  17: 1-65, 2016.

\bibitem{Guo2019}
Z. C. Guo, L. Shi, and S. B. Lin. Realizing data features by deep nets, IEEE Tran. Neural Netw Learn. Syst., In Press. (arXiv: 1901.00130).


\bibitem{Gyorfi2002}
L. Gy\"{o}rfy, M. Kohler, A. Krzyzak and H. Walk. A
Distribution-Free Theory of Nonparametric Regression. Springer,
Berlin, 2002.

\bibitem{Han2019}
Z. Han, S. Yu, S. B. Lin, and  D. X. Zhou.  Depth-selection for deep ReLU nets in feature extraction and generalization.  IEEE Trans. Pattern Anal. Mach. Intel., Revised, 2019.

%
%
%
%

\bibitem{Hinton2006}
G. E. Hinton, S. Osindero, and Y. W. Teh. A fast learning algorithm for
deep belief netws. Neural Comput., 18: 1527-1554, 2006.

\bibitem{Hou2012}
X. Hou, J. Harel, and C. Koch. Image signature: Highlighting sparse salient regions. IEEE Trans. Pattern Anal. Mach. Intel., 34: 194-201, 2012.


\bibitem{Ismailov2014}
V. E. Ismailov. On the approximation by neural networks with bounded
number of neurons in hidden layers. J. Math. Anal. Appl., 417:
  963-969, 2014.


\bibitem{Kohler2014}
M. Kohler. Optimal global rates of convergence for noiseless
regression estimation problems with adaptively chosen design. J.
Multivariate Anal., 132: 197-208, 2014.



\bibitem{Kohler2017}
M. Kohler and A. Krzyzak. Nonparametric regression based on
hierarchical interaction models.  IEEE Trans. Inform. Theory, 63: 1620-1630, 2017.


\bibitem{Krizhevsky2012}
A. Krizhevsky, I. Sutskever, and G. E. Hinton. Imagenet classification
with deep convolutional neural networks. NIPS, 2097-1105, 2012.


%



\bibitem{Lee2009}
H. Lee, P. Pham, Y. Largman, and A. Y. Ng. Unsupervised feature learning
for audio classification using convolutional deep belief networks.
NIPS, 469-477, 2010.

\bibitem{LinH2017}
H. W. Lin, M. Tegmark, and D. Rolnick. Why does deep and cheap
learning works so well? J. Stat. Phys.,   168: 1223-1247, 2017.


\bibitem{Lin2017JMLR}
S.~B. Lin, X.~Guo, and D.~X. Zhou.
 Distributed learning with regularized least squares.
 J. Mach. Learn. Res., 18 (92):  1--31, 2017.


\bibitem{Lin2017a}
S. B. Lin. Limitations of shallow nets approximation. Neural
Networks,   94:  96-102, 2017.


\bibitem{Lin2018CA}
S. B. Lin and D. X. Zhou. Distributed kernel-based gradient descent
algorithms. Constr. Approx., 47: 249-276, 2018.

\bibitem{Lin2018}
S. B. Lin. Generalization and expressivity for deep nets. IEEE
Trans. Neural Netw. Learn. Syst., 30: 1392-1406, 2019.

\bibitem{McCane2017}
B. McCane and L. Szymanski. Deep radial kernel networks: approximating
radially symmetric functions with deep networks. arXiv preprint
arXiv:1703.03470, 2017.

\bibitem{Maiorov1999b}
V. Maiorov  and A. Pinkus. Lower bounds for approximation by MLP neural
networks. Neurocomputing,  25: 81-91, 1999.

\bibitem{Meister2016}
M. Meister and I. Steinwart.  Optimal Learning Rates for Localized
SVMs. J. Mach. Learn. Res.,  17: 1-44, 2016.


\bibitem{Mhaskar1996}
H. N. Mhaskar. Approximation properties of a multilayered feedforward artificial neural network.  Adv. Comput. Math.,   1: 61-80, 1993.


\bibitem{Mhaskar2016a}
H. N. Mhaskar and T. Poggio. Deep vs. shallow networks: An
approximation theory perspective, Anal. Appl., vol. 14, pp. 829-848,
2016.


%

\bibitem{Safran2016}
I. Safran and O. Shamir. Dept-width tradeoffs in approximating
natural functions with neural networks. arXiv reprint
arXiv:1610.09887v2, 2016.

\bibitem{Petersen2017}
P. Petersen and F. Voigtlaender. Optimal approximation of piecewise
smooth functions using deep ReLU neural networks. Neural Networks,
  108: 296-330, 2018.

\bibitem{Schwab2018}
C. Schwab and J. Zech. Deep learning in high dimension: Neural network
expression rates for generalized polynomial chaos expansions in UQ.
Anal. Appl.,   17:  19-55, 2019.

\bibitem{Shaham2015}
U. Shaham,  A. Cloninger, and R. R. Coifman. Provable approximation
properties for deep neural networks. Appl. Comput. Harmon. Anal.,  44: 537-557, 2018.

\bibitem{Shannon1949}
C. E. Shannon. Communication in the presence of noise. Proc. Inst.  Radio Enginer. 37: 10-21, 1949.

\bibitem{Silver2016}
D. Silver, A. Huang, C. J. Maddison, A. Guez, L. Sifre,  G.
van den Driessche, J. Schrittwieser, I. Antonoglou, V.
Panneershelvam, M. Lanctot, et al. Mastering the
game of Go with deep neural networks and tree search.
Nature, 529(7587):  484-489, 2016.


\bibitem{Wright2010}
J. Wright, Y. Ma, J. Mairal, G. Sapiro, T. S. Huang, and S.  Yan.
Sparse representation for computer vision and pattern recognition.
Proc. IEEE,  98: 1031-1044, 2010.

\bibitem{Wu2005}
Q. Wu and D. X. Zhou. SVM soft margin classifiers: linear programming
versus quadratic programming.  Neural Comput.,  17:
1160-1187, 2015.


\bibitem{Yang2009}
J. Yang, K. Yu, Y. Gong, and T. S. Huang. Linear spatial pyramid matching using sparse coding for image classification.   CVPR   1 (2):   6-13, 2009.

\bibitem{Yarotsky2017}
D. Yarotsky. Error bounds for aproximations with deep ReLU networks.
Neural Networks,   94: 103-114, 2017.

\bibitem{Zayed2018}
A. I. Zayed. Advances in Shannon's sampling theory. Routledge, 2018.


%


\bibitem{Zhou2006}
{D. X. Zhou and K. Jetter. Approximation with polynomial kernels and
SVM classifiers, Adv. Comput. Math., 25: 323-344, 2006.}



\bibitem{Zhou2018}
D. X. Zhou. Deep distributed convolutional neural networks:
Universality. Anal. Appl., 16: 895-919, 2018.

\bibitem{Zhou2018a}
D. X. Zhou. Universality of Deep Convolutional Neural Networks. Appl. Comput. Harmonic Anal., DOI: 10.1016/j.acha.2019.06.004
 (arXiv:1805.10769).

\bibitem{Zhou2019}
D. X. Zhou. Theory of convolutional neural networks: downsampling. Neural Networks, Minor Revision, 2019.

\bibitem{Zhou2014}
Z. H. Zhou, N. V. Chawla, Y. Jin, and G. J. Williams. Big data
opportunities and challenges: Discussions from data analytics
perspectives. IEEE Comput. Intel. Mag., 9: 62-74, 2014.




\end{thebibliography}
\end{document}